%% file: main.tex
\documentclass[10pt,twocolumn,letterpaper]{article}

\usepackage{iccv}              


\definecolor{iccvblue}{rgb}{0.21,0.49,0.74}
\usepackage[pagebackref,breaklinks,colorlinks,allcolors=iccvblue]{hyperref}
\usepackage{url}
\usepackage{graphicx} 
\usepackage[capitalize]{cleveref}
\usepackage{booktabs}
\usepackage[inline]{enumitem}
\usepackage{colortbl}
\usepackage{subcaption}
\usepackage{caption}
\usepackage{placeins}
\definecolor{color1}{gray}{0.95} 
\definecolor{color2}{gray}{0.9}  
\definecolor{color3}{gray}{0.85} 
\newcommand{\itb}{\texttt{RadioActive}}
\newcommand{\NDatasets}{ten}
\def\paperID{10997} 
\def\confName{ICCV}
\def\confYear{2025}

\title{RadioActive: 3D Radiological Interactive Segmentation Benchmark}

\author{Constantin Ulrich \thanks{Equal contribution. Authors are permitted to list their name first in their CVs.}$^{~~1,4}$, \quad Tassilo Wald$^{~*1,2,3}$, \quad Emily Tempus $^{*1}$, \\ Maximilian Rokuss$^{1,3}$, \quad Paul Jaeger$^{5,6}$, \quad Klaus Maier-Hein$^{1,2,3,4,5,6}$\\
\small $^{1}$~Division of Medical Image Computing, German Cancer Research Center (DKFZ); \\ \small $^{2}$~Helmholtz Imaging, DKFZ;
\small$^{3}$~Faculty of Mathematics and Computer Science, University of Heidelberg; \\
\small$^{4}$~Medical Faculty Heidelberg, University of Heidelberg;
\small$^{5}$~Interactive Machine Learning Group, DKFZ Heidelberg;\\
\small$^{6}$~Pattern Analysis and Learning Group, Department of Radiation Oncology \\ 
{\tt\small constantin.Ulrich@dkfz-heidelberg.de}
}

\begin{document}

\maketitle
\input{sec/0_abstract}    
\input{sec/1_intro}
\input{sec/2_framework}

\input{sec/3_experiments}
\input{sec/4_summary}
{
    \small
    \bibliographystyle{ieeenat_fullname}
    \bibliography{main}
}

\input{sec/X_suppl}

\end{document}

%% file: sec/0_abstract.tex
\begin{abstract}
Effortless and precise segmentation with minimal clinician effort could greatly streamline clinical workflows.
Recent interactive segmentation models, inspired by META’s Segment Anything, have made significant progress but face critical limitations in 3D radiology. These include impractical human interaction requirements such as slice-by-slice operations for 2D models on 3D data and a lack of iterative refinement. Prior studies have been hindered by inadequate evaluation protocols, resulting in unreliable performance assessments and inconsistent findings across studies.
The \itb~benchmark addresses these challenges by providing a rigorous and reproducible evaluation framework for interactive segmentation methods in clinically relevant scenarios. It features diverse datasets, a wide range of target structures, and the most impactful 2D and 3D interactive segmentation methods, all within a flexible and extensible codebase.  We also introduce advanced prompting techniques that reduce interaction steps, enabling fair comparisons between 2D and 3D models. Surprisingly, SAM 2 outperforms all specialized medical 2D and 3D models in a setting requiring only a few interactions to generate prompts for a 3D volume. This challenges prevailing assumptions and demonstrates that general-purpose models surpass specialized medical approaches.
By open-sourcing \href{https://github.com/MIC-DKFZ/radioactive}{RadioActive}, we invite researchers to integrate their models and prompting techniques, ensuring continuous and transparent evaluation of 3D medical interactive models. 
\end{abstract}

%% file: sec/1_intro.tex
\section{Introduction}
\label{sec:intro}

\begin{figure}[t]
    \centering
    \includegraphics[width=\linewidth]{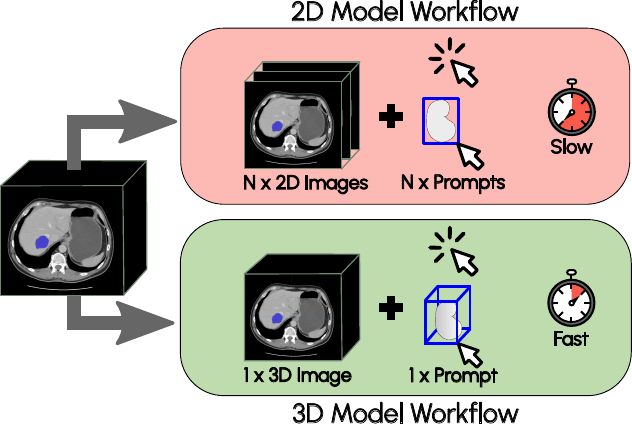}
    \caption{Current interactive segmentation methods require clinicians to interact with radiological images slice-by-slice, leading to increased workload.}
    \label{fig:motivation}
\end{figure}
\begin{figure*}[t]
    \centering
    \includegraphics[width=\linewidth]{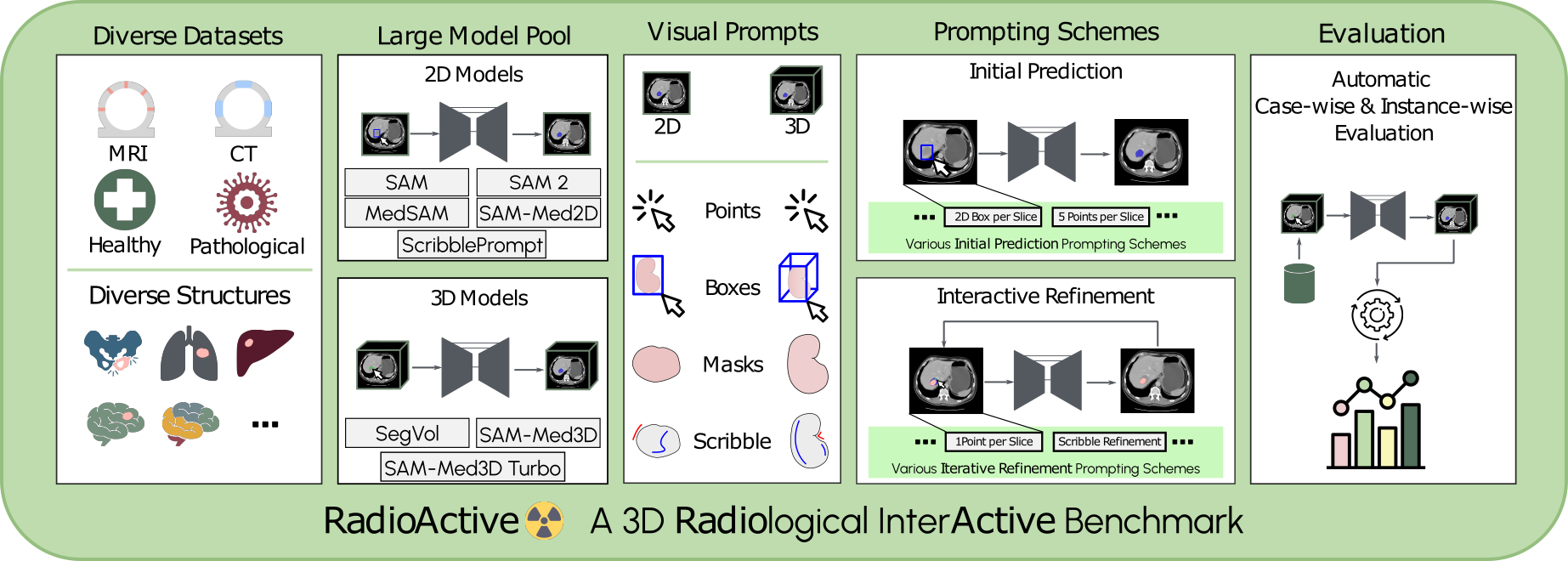}
    \caption{\itb~overview. Although our evaluation is performed on entire 3D volumes, the benchmark accommodates both 3D and 2D interactive segmentation methods. While 3D model prompting is relatively straightforward, we introduce prompting and refinement strategies for 2D models that minimize the effort required from human interaction. The benchmark is designed to be extensible, and researchers are encouraged to propose and integrate additional methods seamlessly using our codebase, particularly for areas marked by three dots.}
    \label{fig:benchmark}
\end{figure*}
Accurate segmentation of anatomical structures or pathological areas is crucial in fields like radiology, oncology, and surgery to isolate affected regions, monitor disease progression, and treatment planning, and guide therapeutic procedures. On the one hand, traditional supervised medical segmentation models have demonstrated strong performance across a range of anatomies and pathologies \citep{nnunet,extending,huang2023stu,ulrich2023multitalent}. However, their applicability remains heavily constrained by the limited task-specific training data available, as well as their deterministic predictive behavior that cannot be influenced, possibly requiring labor-intensive manual post-processing of predictions. Consequently, fully autonomous AI solutions have not yet found widespread autonomous clinical applications. 
\\
On the other hand, numerous semi-automatic segmentation techniques, not reliant on AI, are already in clinical practice to expedite manual annotation processes \cite{Hemalatha18}. These current ad hoc methods do not yet tap into the potential of AI-based automation to drastically reduce annotation time. A method that bridges this gap and allows clinicians to segment any target with just a single click within the image could greatly enhance the efficiency of clinical workflows. 
\\
Interactive segmentation methods, like the Segment Anything Model (SAM) \citep{sam} represent exactly this. SAM is designed to segment any target through various user interaction methods, including point-based and bounding box prompts. This allows users to easily specify the area of interest by clicking on it or drawing a bounding box around it, making the segmentation process both flexible and intuitive. 
A particularly powerful feature is the ability for users to iteratively refine initial predictions by adding more positive or negative prompts. 
\\
Due to their great potential, interactive segmentation methods have attracted a lot of attention in the medical domain and led to many studies evaluating and adapting SAM for 3D medical image segmentation \citep{sammd, samdigital, sammeetsmedicalimages, samsegmentpolyps, samvsbet, sammed2d, medsam, 3dsamadapter, ScribblePrompt}. Moreover, several researchers have been inspired by SAM's capabilities to develop their own methods, often specifically designed for 3D medical data \citep{segvol,vista3d,prism,sammed3d}. 
\\
Although these domain-specific adaptations to medical data have shown promising progress, many published methods are plagued by pitfalls that obfuscate the efficacy of the models and prevent clinicians and researchers from determining the best methods for their use cases:
\\
\textbf{Applying interactive 2D models to 3D data on a slice-by-slice basis (P1):} Assuming clinicians will interact with each slice individually is unrealistic and undermines the efficiency improvements these methods aim for. Moreover, a slice-by-slice approach introduces an unfair bias when comparing 2D and 3D models, as 3D models typically require only a few interactions per image, leading to significantly fewer interactions and less supervision \cite{sammed2d, medsam, SAMed, OnePrompt,ScribblePrompt, mazurowski2023segment}.
\\
\textbf{Neglecting refinement (P2)}: 
Many studies assess interactive segmentation methods based on a single interaction step, overlooking the inherent ambiguities in radiological images \citep{medsam,segvol,3dsamadapter,sam3d}.
Often, a second interaction may be necessary to specify which specific substructure the clinician wants to segment. This could be, e.g. a vessel within the liver, or the necrosis within a tumor, as exemplified in the well-known BraTs segmentation challenge \citep{brats}.
Furthermore, clinicians often want to adapt the segmentations to their clinic's local protocol or refine them, particularly for targets with high inter-rater variability, like pathological structures \citep{interraterspine, interobserverradiology, deeplearningahievements}.
Overall, there is a notable lack of research exploring realistic, iterative refinement methods for 2D models applied to 3D volumes.
\\
\textbf{Obfuscated and insufficient evaluation (P3):}Currently, no standardized evaluation framework exists, leading to inconsistencies across studies. As a result, comparisons between papers are often unreliable, with evaluations frequently being opaque or insufficient.
We observed the following shortcomings:
\begin{enumerate*}[label=(\roman*)]
    \item Not specifying whether predictions were interactively refined or based on a single prompt with multiple points \citep{sammed2d,sammed3d}.
    \item Being intransparent on the number of initial prompts given \citep{segvol}.
    \item Using the best mask rather than the final mask after interactive refinement \citep{sammed3d}.
    \item Evaluating predictions slice-by-slice or on sub-patches of a 3D volume instead of evaluating the full image \citep{sammd, medsam, sammed2d, vista3d, prism}.
    \item Excluding targets considered \textit{too small}, hence neglecting valid targets such as small lesions that are neither tested nor trained on \cite{medsam, sammed2d, sammed3d}.
    \item Comparing against non-promptable models and SAM, rather than any other promptable model trained on medical data \citep{sammed2d, medsam, 3dsamadapter, vista3d}.
    \item Lastly, overemphasizing segmenting healthy structures, such as organs, where existing supervised public models already perform well \citep{wasserthal2023totalsegmentator,ulrich2023multitalent}, instead of focusing on pathologies, where interactive refinement could provide the greatest benefits \citep{sammed3d, SAMed}.
\end{enumerate*}
\\
Developing a benchmark is key to overcoming these challenges, in line with the recent recommendations by \citet{Marinov}. To this end, we present \itb, a robust, reproducible, and extendable \textbf{Radio}logical Inter\textbf{Active} Benchmark for 3D medical image segmentation. This benchmark highlights the most effective 2D and 3D interactive segmentation approaches and tests realistic prompting methods in the 3D domain. We conduct evaluations directly in the 3D domain and across multiple models and datasets. Our key contributions include: 
\begin{enumerate}
\item  \itb, for the first time, enables a fair comparison of the most influential 2D and 3D interactive segmentation methods for 3D medical data. By measuring the number of simulated interactions, a proxy for the 'Human Effort', we test different prompting strategies that do not require slice-by-slice interactions (P1).
\item  We propose effective interaction strategies for refinement of predictions in a 3D volume, without requiring clinicians to interact with each individual slice (P2).
\item We provide a standardized evaluation protocol to generate prompts, select model outputs and compute the segmentation metrics on the entire image across \NDatasets~datasets, covering various modalities and target structures, including small lesions (P3). Our benchmarking efforts include a performance comparison against leading interactive segmentation methods in the medical domain.
\item The extendable \itb~framework allows developers to a) easily evaluate a new method in a fair manner against established methods and b) easily develop and investigate new prompting strategies.
\end{enumerate}
\noindent
Through open-sourcing \itb, we invite researchers to integrate their methods into our framework, promoting continuous and equitable assessment that allows reproducibility and transparently tracking the overall progress in interactive 3D medical image segmentation.   

\begin{figure}[t]
    \centering
    \includegraphics[width=\linewidth]{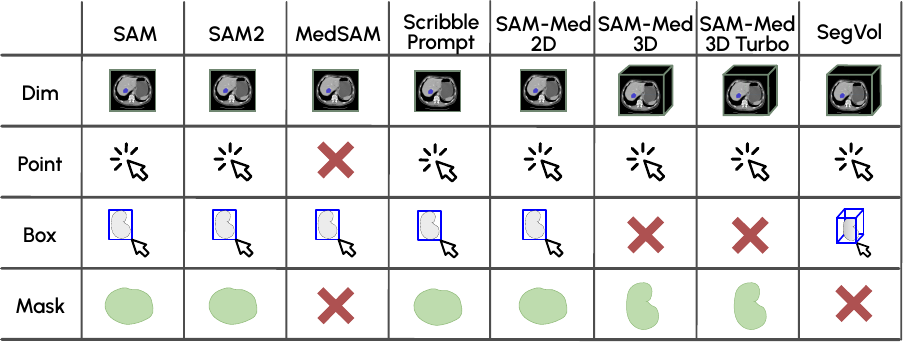}
    \caption{Some models operate natively in 3D and enable full 3D interaction. Only models that accept mask prompts allow iterative refinement of initial predictions with human guidance.}
    \label{fig:overview}
\end{figure}

%% file: sec/2_framework.tex
\section{RadioActive}

The \itb~Benchmark is designed to easily enable a fair and reproducible evaluation of 2D and 3D interactive segmentation methods for 3D radiological image segmentation for the very first time \cref{fig:benchmark}.
 In \cref{apx:sec:literature_review}, we include an in-depth overview of the state of current independent benchmarkings for interactive segmentation methods for 3D radiological images \cite{review_sam, Marinov, sammd, mazurowski2023segment, SAMBenchmarkHe, IMIS, zhang2024unleashingpotentialsam2biomedical, antonov2024rclicksrealisticclicksimulation}.
While prompting 3D models is generally straightforward, we introduce specific prompting and refinement strategies for 2D models to streamline human interaction and reduce the simulated \textit{Human Effort}. The proposed benchmark includes seven established models and \NDatasets~datasets covering different target structures and image modalities. All datasets are publicly available and we provide an automatic download and pre-processing for improved usability and reproducibility.
Moreover, the benchmark is built with flexibility in mind, enabling seamless integration of additional methods. Researchers are invited to contribute new approaches, particularly new models, new prompting schemes, and new interesting datasets to the collection. Overall, the design of our benchmark allows for easy testing and validation of novel segmentation methods, making the benchmark a catalyst for advancing methodology for interactive 3D medical image segmentation. In the following, we present the different components of \itb. 

\label{sec:Methods}
\subsection{Initial Prompting}\label{subsec:InitialPrompting}
\begin{figure}[b]
    \centering
    \includegraphics[width=0.99\linewidth]{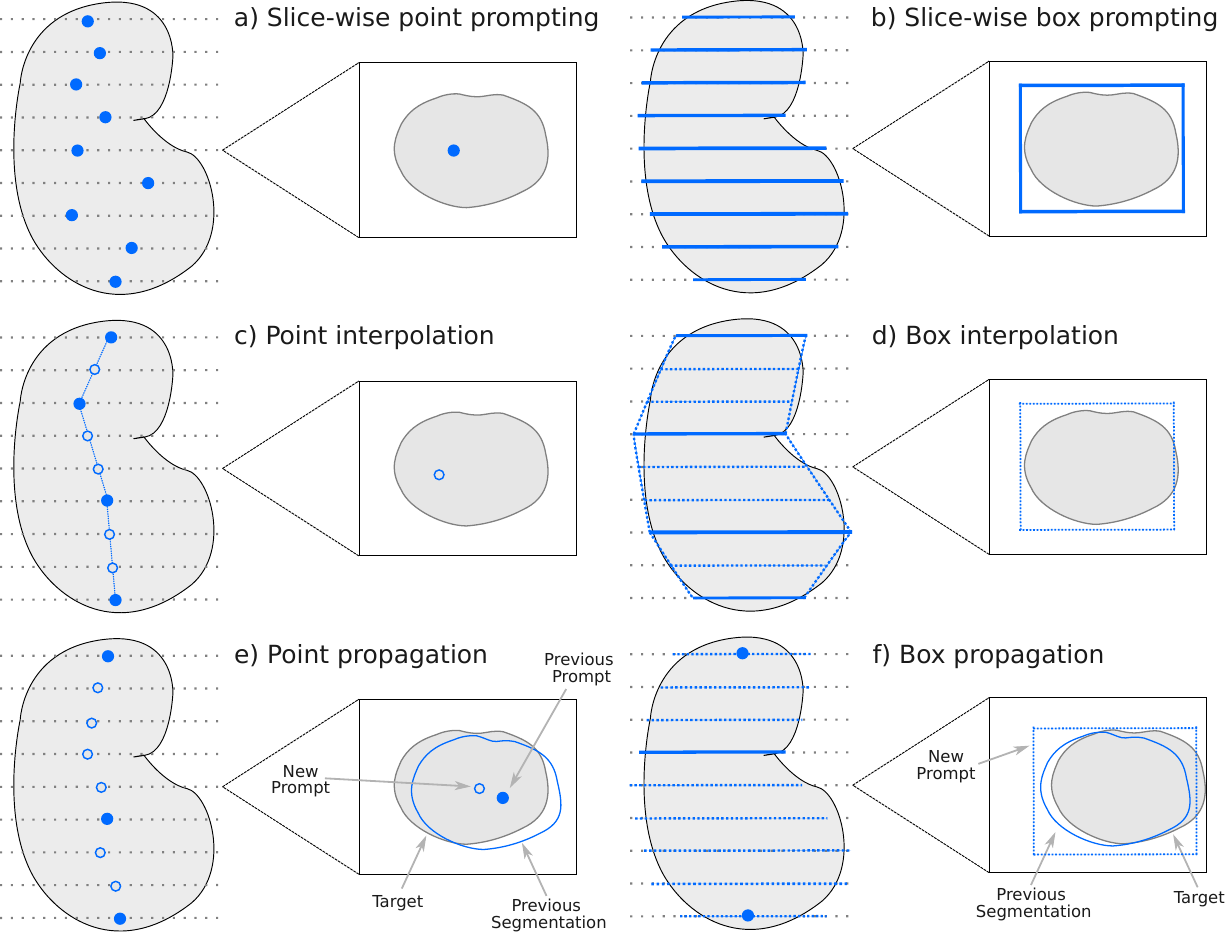}
    \caption{Different prompting schemes for 2D models based on point prompts (on the left) and box prompts (on the right). While a) and b) expect unrealistic human slice-by-slice interaction, c) and d) illustrate the proposed prompt interpolation schemes, where a human needs to provide prompts for at least 3 slices (4 slices in this case). Prompts for the remaining slices are generated by interpolating between the initial prompts.  e) and f) present the proposed prompt propagation methods, where the prompt for each subsequent slice is automatically generated based on the model's prediction from the previous slice. Only the initial slice and upper and lower boundaries require manual prompts.}
    \label{fig:prompting}
\end{figure}

Prompts are a key component of any interactive segmentation method and can highly influence the achieved performance of the underlying method. \itb~distinguishes between two visual prompting types. Point prompts correspond to a click of a user in the image, and box prompts refer to a box around the target structure. Notably, some methods also enable a distinction between foreground and background point prompts.
While 3D models allow segmenting a 3D volume natively, 2D-based models require an interaction for each slice, resulting in excessive effort, which is prohibitive for clinicians as it would take too much time in daily clinical practice.
Hence, any meaningful benchmark must account for this difference in prompting effort. 
To increase the feasibility of 2D models for 3D applications, it is essential to reduce this effort. We propose two straightforward methods, for both point and box prompts, to explore their performance and provide a proxy for measuring the effort of human interaction. 
\\
\textbf{Point interpolation}: Let $I \subseteq N$ be the set of axial indices of all foreground slices. We simulate a user by selecting $n$ foreground points, specifically the center of the largest connected component of slice $i_1,..., i_n \in I$ where the $i_j$ are equally spaced within $I$ and $i_1 = \min(I)$ and $i_n = \max(I)$. Then, we interpolate linearly between each point and the next one and use the intersections of the resulting lines with the axial slices as positive point prompts, as visualized in \cref{fig:prompting} c). 
\\
\textbf{Point propagation}: We simulate a user providing  \(\min(I)\), \(\max(I)\), and a 2D point prompt within the slice corresponding to the median axial index \(i_{m}\). Given this point, the model generates a segmentation \( S_m \) for the median slice.
We then calculate a \textit{central point}, specifically the center of mass of the largest connected component of \( S_m \), to use as a point prompt for the slice indexed by \(i_{m-1}\).
Again, we generate a segmentation \(S_{m-1}\) of this slice, and create a new central point until we segment the slice with the axial index \(\min(I)\). The propagation is then repeated upwards, starting from \(i_{m+1}\) and continuing until we segment the slice with the axial index \(\max(I)\). This process is visualized in \cref{fig:prompting} e).
\\
\textbf{Box interpolation}: We simulate a user providing $n$ 2D bounding boxes, one in each of $i_1, ...., i_n \in I$, with $i_j$ defined as in the point interpolation paragraph. Since the boxes are uniquely defined by their minimum and maximum vertices, we can interpolate between the minimum vertices as in point propagation to get a minimum vertex in each axial slice, and similarly get a maximum vertex in each axial slice, providing a box prompt in each slice. This box interpolation is exemplified in \cref{fig:prompting} d).
\\
\textbf{Box propagation}: We simulate a user providing \(\min(I)\), \(\max(I)\), and a 2D box prompt within the slice corresponding to the axial index \(i_{m}\), with $m$ defined as in point propagation. The model then generates a segmentation \( S_m \) for the median slice. A new bounding box is created based on \(S_m\) and used as a prompt for the slice indexed by \(i_{m}-1\). The propagation is continued down to \(\min(I)\) and then repeated upwards until \(\max(I)\) as in point propagation, but using box prompts instead of point prompts. 
See \cref{fig:prompting} f) for a visualization.
\\
While realistic prompting behavior is the priority \itb~also supports the previously used unrealistic slice-by-slice prompting styles for completeness. Additional details on prompt schemes are provided in the appendix in \cref{apx:sec:prompt_scheme_details}.

\subsection{Refinement Prompting}\label{subsec:refinementprompting}
Refinement of previous segmentations is an important aspect of interactive segmentation models, as it allows iteratively improving the segmentation until the desired structure is segmented to a user's demands. 
Some interactive segmentation models allow for the refinement of initial segmentations by providing the model with the previous prediction along with a new prompt to correct errors, either through foreground clicks on false negative pixels or background clicks on false positive pixels. While this process is straightforward for 3D models, 2D models naively only allow for refinement in a slice-by-slice fashion, which again places an unrealistic burden on clinicians. Therefore, we present refinement strategies that require a reasonable level of \textit{Human Effort}.

\paragraph{Scribble refinement:} 
To represent a user-centric refinement strategy we introduce an algorithm simulating user-created scribble prompts:
At each refinement step, our proposed algorithm generates either positive or negative additional prompts. The decision to generate positive prompts follows a Bernoulli trial with success probability $p= n_{fn}/(n_{fn}+n_{fp})$, where $n_{fn}$, $n_{fp}$ represent the number of false negatives and false positive voxels, respectively.  \\
If positive prompts are selected, we perform a connected component analysis on the false negative voxels. Given $L$, the largest connected component, we generate a scribble from the bottom to the top of $L$ by taking the centroid of $L$ in each slice to simulate drawing a vertical scribble through the \textit{middle} of $L$. This simulates a clinician annotating regions that were erroneously not segmented. \\
For 2D models, we then individually feed all slices $i \in I$ where the voxel along the scribble was not predicted, along with the new positive prompt derived from the scribble and the previous prediction $s \subset S$, back into the model. 
For 3D models, we feed the whole 3D patch, together with the previous prediction $S$, and a random point derived from the scribble into the network in one step.  \\
If negative prompts are selected, we identify a non-axial slice $S_{fp}$ of $S$ that contains the most false positives. Then we generate a contour curve around the ground truth target object at a distance of 2 pixels. We then select a subpart $C$ with a length of 60\% of the full curve and sample all pixels $c \in C$ that are false positives to obtain a set of points $D$, simulating a user drawing a few scribbles in areas where the model over-segmented the target. 
For 2D models, we then generate new slice predictions for each slice containing a point in $D$ by providing the model with the previous prediction as well as new negative prompts: all $d\in D$ which belong to that slice. For 3D models, we again feed the whole 3D patch, together with the previous prediction $S$ and a negative prompt sampled from $D$. The motivation for the different approaches for FPs is, that correcting the model near the target boundary is likely to provide more informative feedback. A visualization of the proposed scribble refinement is provided in the appendix in \cref{fig:prompting_refine}.
As none of the evaluated methods features a dedicated scribble prompt encoding, we instead generate a set of point prompts from our scribble refinement trajectory, which is similar to the recently proposed work of \citet{ScribblePrompt} for scribble prompting of SAM.
Aside from this refinement strategy, we also support unrealistic refinement prompts for which details are provided in the appendix in \cref{apx:sec:prompt_scheme_details}.

\subsection{Human Effort Proxy}
A model's performance is highly dependent on the effort a human invests into the initial prompting and the refinement of the predicted masks. Generally, the effort of using 3D methods is lower than the effort of using 2D methods, although the strategies mentioned above significantly reduce the effort of 2D methods.

\noindent
We aimed to establish a quantitative measure of the effort a method would require from a human user, however, various issues arise:
Considering a formalized mathematical approach where effort is quantified through the degrees of freedom (DoF) of each interaction. A point would correspond to 3 DoF, a 2D box 5 DoF (requiring selection of the z-axis and two 2D points), and a 3D box corresponds to 6 DoF. The point interpolation corresponds to 9 DoF, whereas point propagation would only be 5 DoF, since it requires just the axial coordinate rather than both minimum and maximum points with 3 DoF each. From the user’s perspective, however, identifying the z-coordinate demands the same level of effort as selecting a 3D coordinate by clicking at the target structure’s endpoint along the z-axis. Similarly, an arbitrary scribble has significantly more DoF than a straight or parabolic line, yet the difference in effort for the user is minimal.
Subsequently, this approach was discarded.

\noindent
Instead, we define user effort in terms of the \textbf{number of interactions} (i.e. clicks of a user) required for a specific prompt. This leads to e.g. a 2D and 3D click being equivalent in cost. While this is not an exact measure, it offers the most pragmatic estimation of the actual effort involved from a user’s perspective.

\subsection{Interactive Segmentation methods}
In our benchmark, we include various interactive segmentation methods. \cref{fig:overview} illustrates the types of prompts each method supports. Iterative refinement is only possible for methods that allow a (previously predicted) mask as a prompt.\\
\noindent
\textbf{SAM} is the most prominent model from the natural image domain, that inspired many researchers to evaluate and adapt it to the domain of radiological medical images. It was trained on iteratively generated and curated 1B masks and 11M images, but not explicitly on radiological images. It was the first to popularize interactive segmentation models \citep{sam}. \\
\noindent
\textbf{SAM2} is an extension of SAM that was trained on even more images and introduced support for video data \citep{sam2}. \\
\noindent
\textbf{MedSAM} is an adaptation of SAM that fine-tuned SAM's weights on 1,570,263 image-mask pairs from the medical domain. It supports only a single forward pass without refinement and is limited to box prompts \citep{medsam}.\\
\noindent
\textbf{SAM-Med 2D} is another adaptation of SAM, fine-tuned on 4.6 million images with 19.7 million masks from the medical domain. Unlike MedSAM, it supports points, boxes, and mask prompts, allowing for refinement \citep{sammed2d}.\\
\noindent
\textbf{SAM-Med 3D} incorporates a transformer-based 3D image encoder, 3D prompt encoder, and 3D mask decoder. It was trained from scratch using 22,000 3D images and 143,000 corresponding 3D masks and supports point and mask prompts and also allows for refinement \citep{sammed3d}. Notably, it supports only one point prompt per prediction.\\
\noindent
\textbf{SAM-Med 3D Turbo} is an updated version of SAM-Med 3D trained on a larger dataset collection of 44 datasets for improved performance. It supports the same prompt styles as SAM-Med 3D \citep{sammed3d}.\\
\noindent
\textbf{SegVol} is an interactive 3D segmentation model based on a 3D adaptation of a ViT \citep{dosovitskiy2020image} that was trained on 96K unlabelled CT images and fine-tuned with 6K labeled CT images. It supports points and bounding boxes as spatial prompts but does not allow iterative refinement \citep{segvol}. \\
\noindent
\textbf{ScribblePrompt} is an interactive 2D segmentation model designed specifically for medical imaging \cite{ScribblePrompt}. It supports multiple input prompts, including points, boxes, and masks, while introducing scribbles as a novel interaction method. The model integrates these prompts as an additional channel within a U-Net architecture and is trained  with 54,000 scans.

\noindent
Aside from these models there exist other notable interactive models, such as Vista3D~\citep{vista3d}, 3D Sam Adapter~\citep{3dsamadapter} and Prism~\citep{prism}. However, while being promptable, they are closed-set, i.e. not trained to segment any arbitrary prompted structure. Subsequently, they were not considered for this benchmark. 


\subsection{Datasets}
Dataset selection was a non-trivial problem for this benchmark:  
While models that were originally introduced in the natural image domain rarely see any radiological 3D data, the medical counterparts were often trained on all publicly available datasets that the authors could obtain.
For example, MedSAM was trained using more than 60 publicly available datasets \citep{medsam}.
Although these methods conducted their final validation on excluded datasets or at least on separate test subsets of images, the test datasets vary between models.
As a result, identifying annotated datasets with interesting target structures that were not part of any of the included methods' training datasets has proven challenging. \\
Nevertheless, we assembled a diverse collection of ten lesser-known or recently released public datasets featuring various pathologies and organs, including CT and MRI image modalities. Specific details of these are provided in \cref{tab:datasets}. While the quantity of cases may seem low, many contain multiple instances or target structures. Moreover, due to the pathological nature of many they represent the currently common size of datasets in the medical domain. Overall, these 10 datasets offer a robust foundation for evaluating interactive segmentation methods, ensuring diverse, clinically relevant challenges while mitigating biases from training data overlap. To enhance reproducibility and eliminate barriers of entry for non-domain experts, we automated the dataset download and pre-processing, minimizing any required domain knowledge to use the benchmark. However, due to the sparsity of labeled datasets, we urge developers to exclude these datasets from their train dataset selection, as inclusion would compromise the integrity of a clean evaluation through \itb.


\begin{table}
\caption{Overview over all Datasets. None of these datasets were part of the original training data for the methods, except for SegVol, which utilized CT images from D2 HanSeg .}
\label{tab:datasets}
\resizebox{\linewidth}{!}{
\begin{tabular}[]{lllr}
\toprule
\textbf{Dataset}            & \textbf{Modality} & \textbf{Targets}         & \textbf{Images}       \\ \midrule
D1 MS Lesion \citep{MUSLIM2022108139}            & MRI (T2 Flair)                & MS Lesions     & 60 \\
D2 HanSeg   \citep{podobnik2023han}   & MR (T1)      &  30 Organs at risk  & 42               \\
D3 HNTSRMFG    \citep{hntsmrg2024wahid}        &    MRI (T2)          & Oropharyngeal cancer \&
metastatic lymph nodes  & 135             \\
D4 RiderLung \citep{rider_lung} & CT & Lung lesions & 58 \\
D5 LNQ    \citep{lnq2023challenge}                 & CT                & Mediastinal lymph nodes & 513\\
D6 LiverMets    \citep{livermets}                 & CT                & Liver metastases & 171 \\
D7 Adrenal ACC    \citep{adrenalacc}                 & CT                &  Adrenal tumors & 53\\
D8 HCC Tace        \citep{hcctace}             & CT                &  Liver, Liver tumors & 65\\
D9 Pengwin    \citep{liu2023pelvic}           & CT                &   Bone fragments          &100      \\
D10 Segrap \citep{luo2023segrap2023}& CT & 45 Organs at risk & 30 \\
\bottomrule
\end{tabular}}
\end{table}

\subsection{Evaluation}
All interactive segmentation methods identify their target structure based on a spatial prompt, inherently resulting in instance segmentation. As a result, we evaluate on an instance-by-instance basis. Unlike in object detection, each prompt already provides information on the localization of the target structure, making detection metrics like F1-Score irrelevant. Subsequently, we rely solely on the Dice Similarity Coefficient (DSC) score as a metric. 
The instance-wise DSC metric is then averaged per case (i.e. per image volume), and further aggregated across all cases in the dataset, as recommended by \citet{metricsreloaded}. For better presentation, we averaged the DSC across all classes of a dataset and also specified how many human interactions were simulated.

%% file: sec/3_experiments.tex
\section{Experiments and Results}
In our experiments, we evaluate all seven models across all datasets. First, we evaluate the performance during the initial prompting step, and following this, we evaluate the iterative refinement performance of methods that support this. For each of these, we test models in both, realistic and high-effort prompting schemes. To enhance readability, we present the results together with the primary findings in the main manuscript, while additional, more granular insights are deferred to \cref{apx:additional_results} in the appendix.

\subsection{Initial prediction}
\begin{figure}[htbp]
    \centering
    \includegraphics[width=\linewidth]{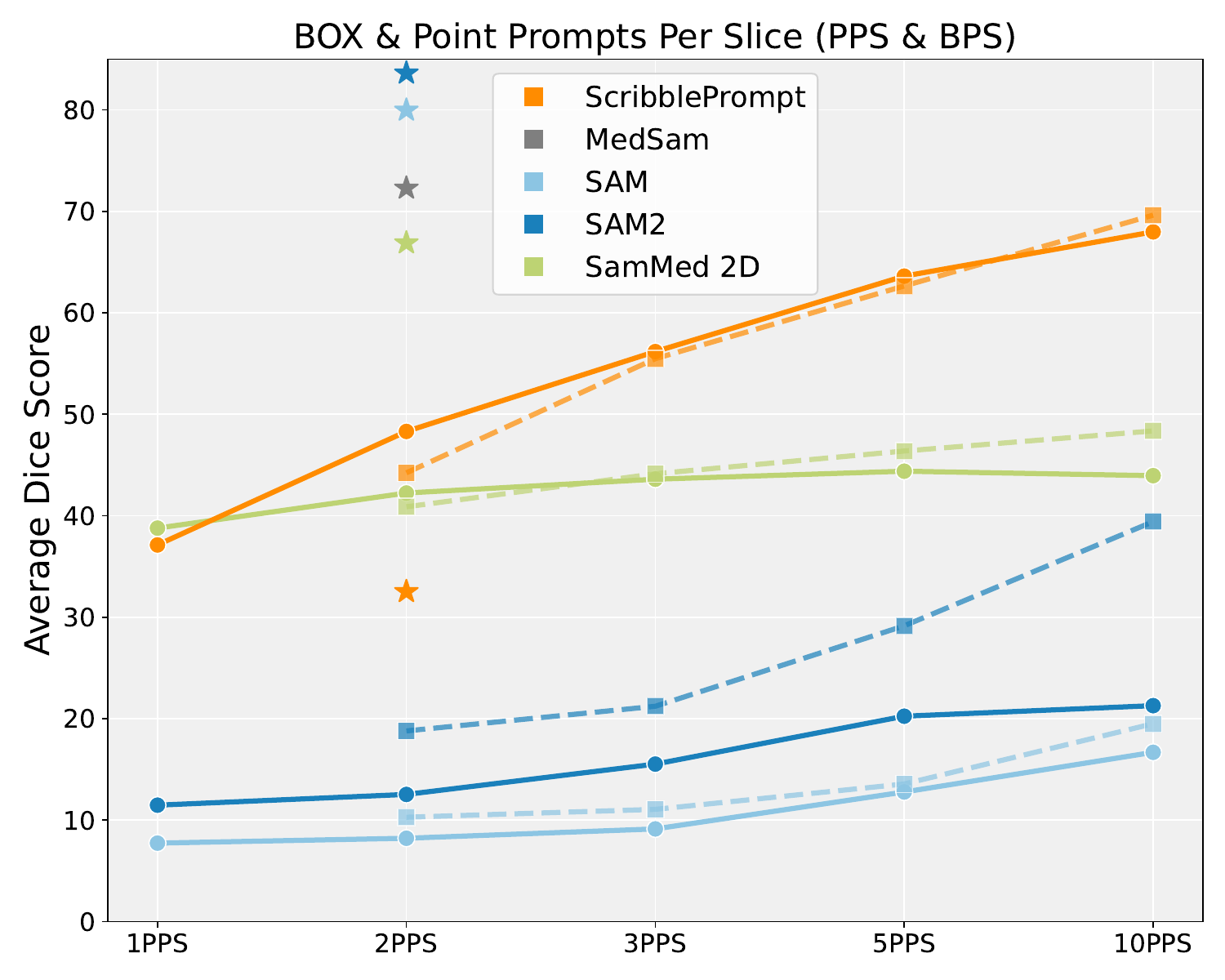}
    \caption{\textbf{Unrealistic prompting of 2D Boxes each slice performs best.} 2D models prompted with one Box Prompt Per Slice (BPS), indicated with a star, outperform models prompted with various Point Prompts per Slice (PPS, line plots). Drawing a 2D box accounts for 2 interactions similar to providing 2 points. Providing alternating positive and negative points (dashed lines) is slightly superior to only positive points.}
    \label{fig:unrealistic}
\end{figure}

\paragraph{Unrealistic effort:} As an upper baseline, we begin with an idealized and unrealistic scenario where each slice is prompted individually for all 2D models.
In this setting, we evaluate different numbers of point prompts per slice (PPS), as well as alternating positive and negative prompts ($\pm$ PPS), and slice-wise box prompts with one box per slice (BPS). \Cref{fig:unrealistic} shows that models employing box prompts achieved significantly higher average Dice scores, with SAM2 demonstrating the strongest performance across all models. Conversely, point-based prompts performed poorly, particularly for small target regions, such as small MS lesions in dataset D1 (see \cref{tab:unrealistic} in the appendix). Although including positive and negative prompts and increasing the number of point prompts led to improvements, only ScribblePrompter achieved performance in the same range as box-based prompts. Interestingly, generalist models not trained with a particular focus on medical data performed worse with point prompts per slice but outperformed models trained on medical data when using box prompts. This may be because models trained with medical data are better at resolving the ambiguity of point prompts, especially given the often low contrast between pathologies and surrounding tissue.  

\begin{figure*}[htbp]
    \centering
    \begin{subfigure}{0.32\linewidth}
        \includegraphics[width=\linewidth]{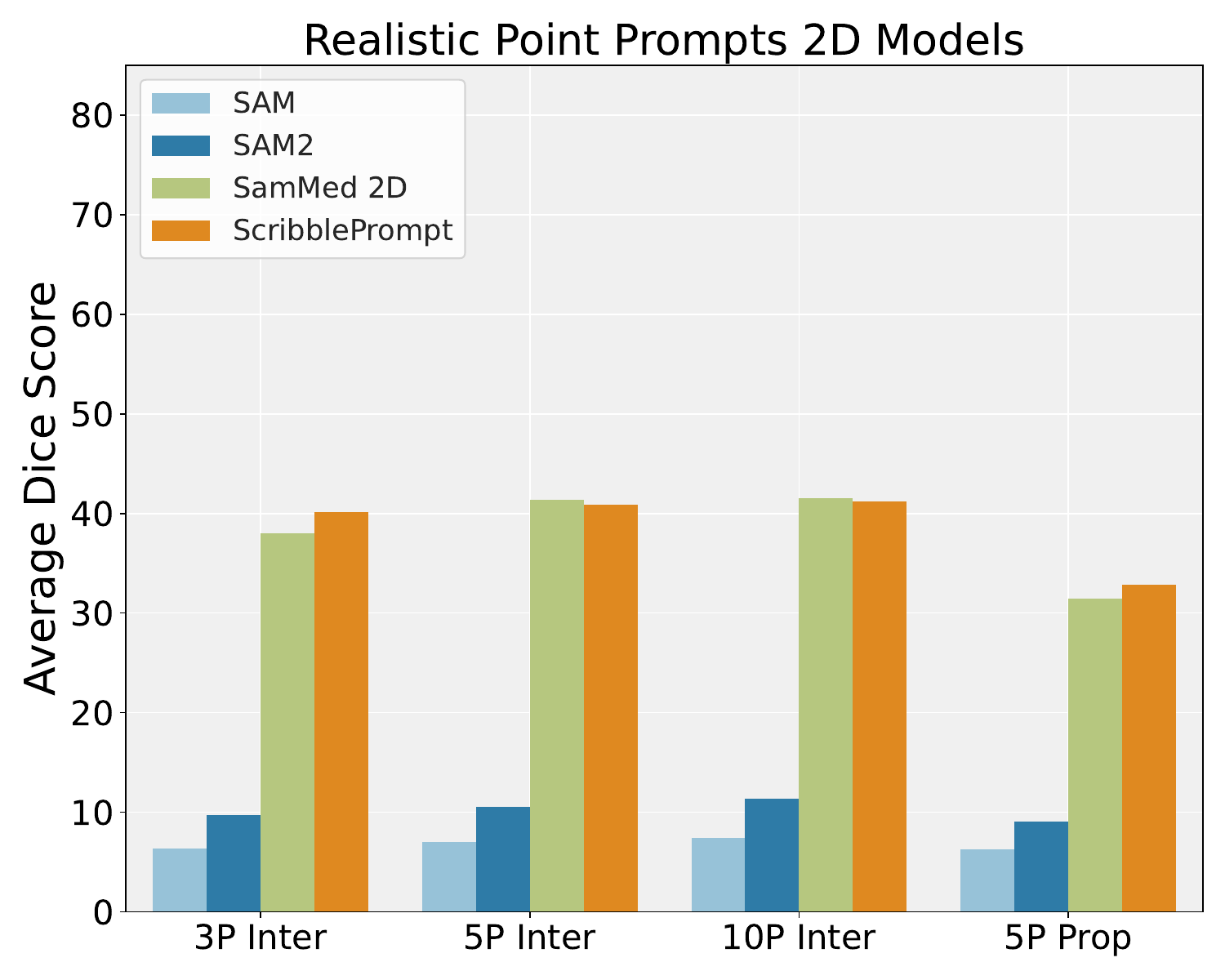}
    \end{subfigure}
    \begin{subfigure}{0.32\linewidth}
        \includegraphics[width=\linewidth]{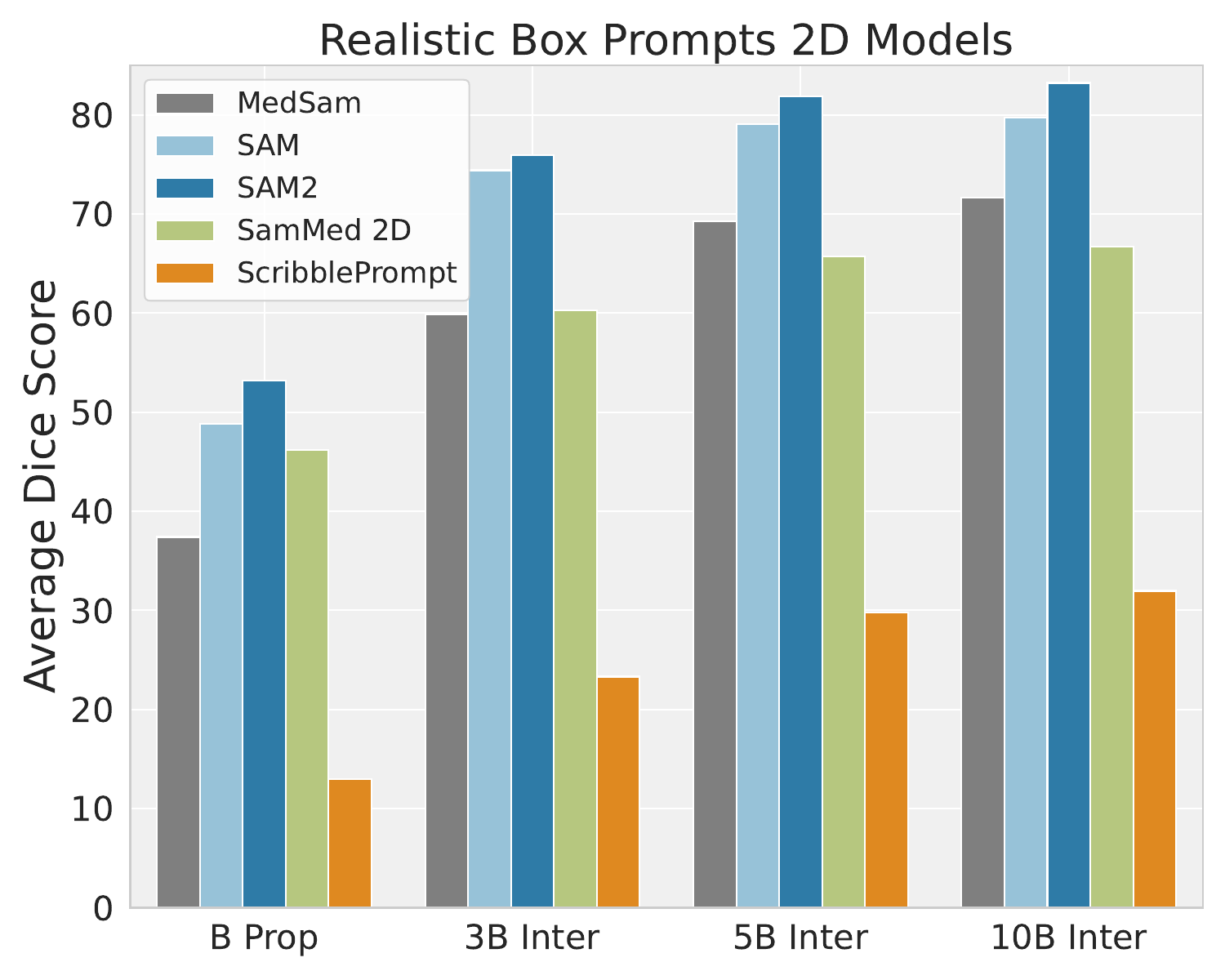}
    \end{subfigure}
        \begin{subfigure}{0.32\linewidth}
        \includegraphics[width=\linewidth]{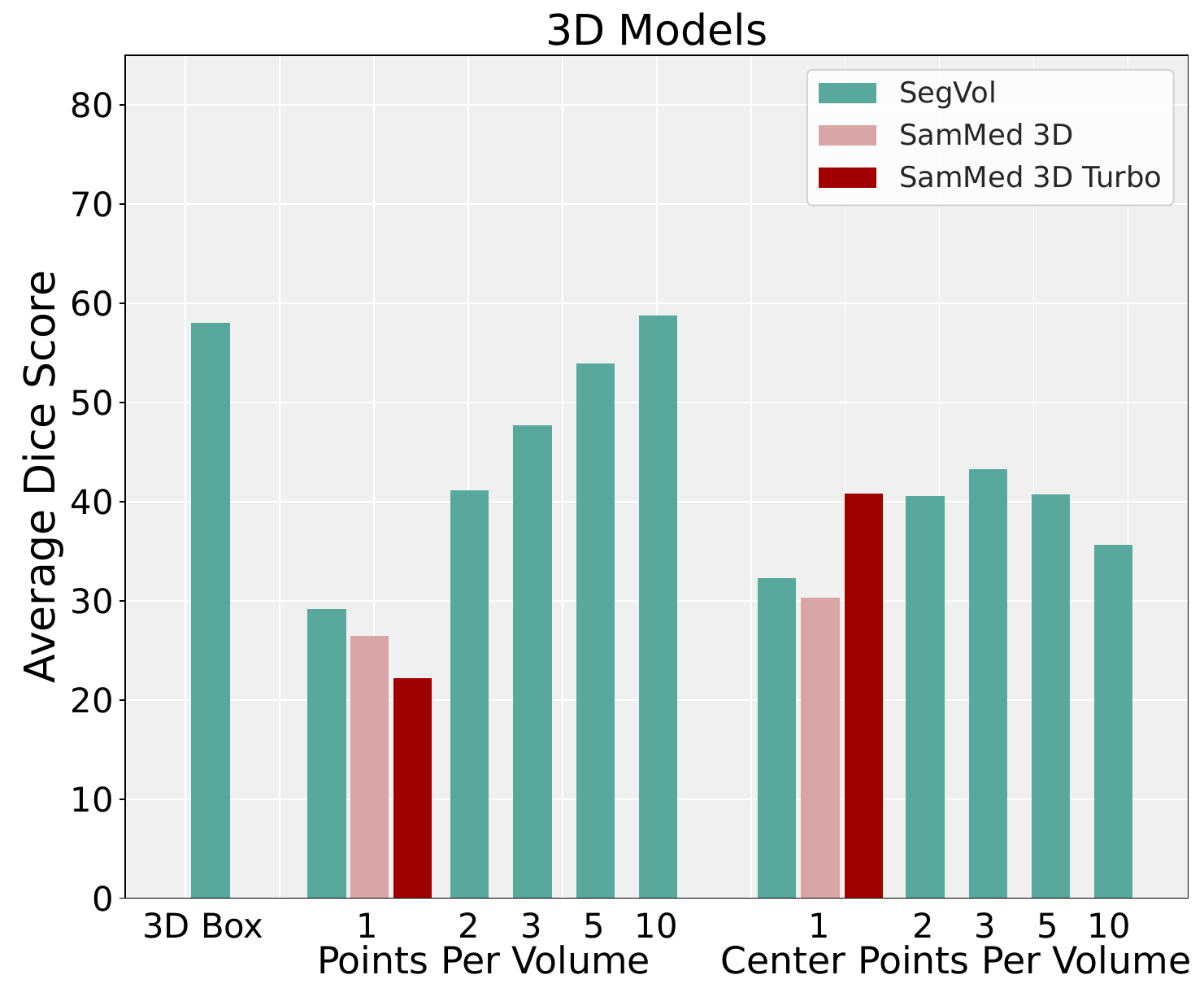}
    \end{subfigure}
    \caption{\textbf{Simple Interpolation Strategies Match Unrealistic Slice-Wise Prompting.} Sampling prompts from the interpolated connection between three initial prompts yields similar performance for SAM2 as slice-wise prompting (left). This is also observed for box interpolation across all models (middle). 3D models perform worse than 2D methods. We compare points sampled at equal axial distances along the five-point interpolation prompter line, which approximates the axial target centerline, with randomly sampled points. Sampling a single point from the center performs better than sampling a random point, but random foreground sampling is better for multiple points.
}
    \label{fig:realistic_static}
\end{figure*}

\paragraph{Realistic Effort:} To simulate a human-in-the-loop scenario, we evaluate various prompting strategies that avoid slice-by-slice interaction. As described in \cref{sec:Methods}, for 2D models, we test point and box interpolation, as well as propagation, using different numbers of initial prompts. For 3D models, we explore varying numbers of Point prompts Per Volume (PPV) and 3D box prompts. \cref{fig:realistic_static} presents the following key findings:

\begin{enumerate}
    \item For all models, simple prompt interpolation methods achieve similar performance as the corresponding prompt per-slice methods. SAM2 achieves a mean Dice of 81.88 when we using 5 initial 2D boxes (5 box inter) compared to a Dice of 83.63 for one box per slice (Box PS), as can be seen in the appendix in \cref{tab:realistic_effort_2d}. 
    \item SAM 2 outperforms specialized medical models across all prompting schemes using box interpolation.
    \item Among 3D models, only SegVol is competitive to 2D models that use box prompts or 5-10 points per volume  as can be seen in the appendix in \cref{tab:realistic_effort_3d}.
    \item Both box and point propagation perform worse than their interpolation counterpart. For propagation schemes, if a previous prompt did not yield a prediction, the propagation would terminate. However, in practice, we observed the opposite effect. Notably, point prompts frequently caused severe oversegmentation, covering the entire body and obscuring the target object, ultimately leading to a loss of meaningful information. Qualitative examples of this failure case are presented in the appendix in \cref{apx:additional_results}. This issue may improve as models continue to evolve.
\end{enumerate}

\subsection{Iteractive Refinement}
\begin{figure}[htbp]
    \centering
        \includegraphics[width=\linewidth]{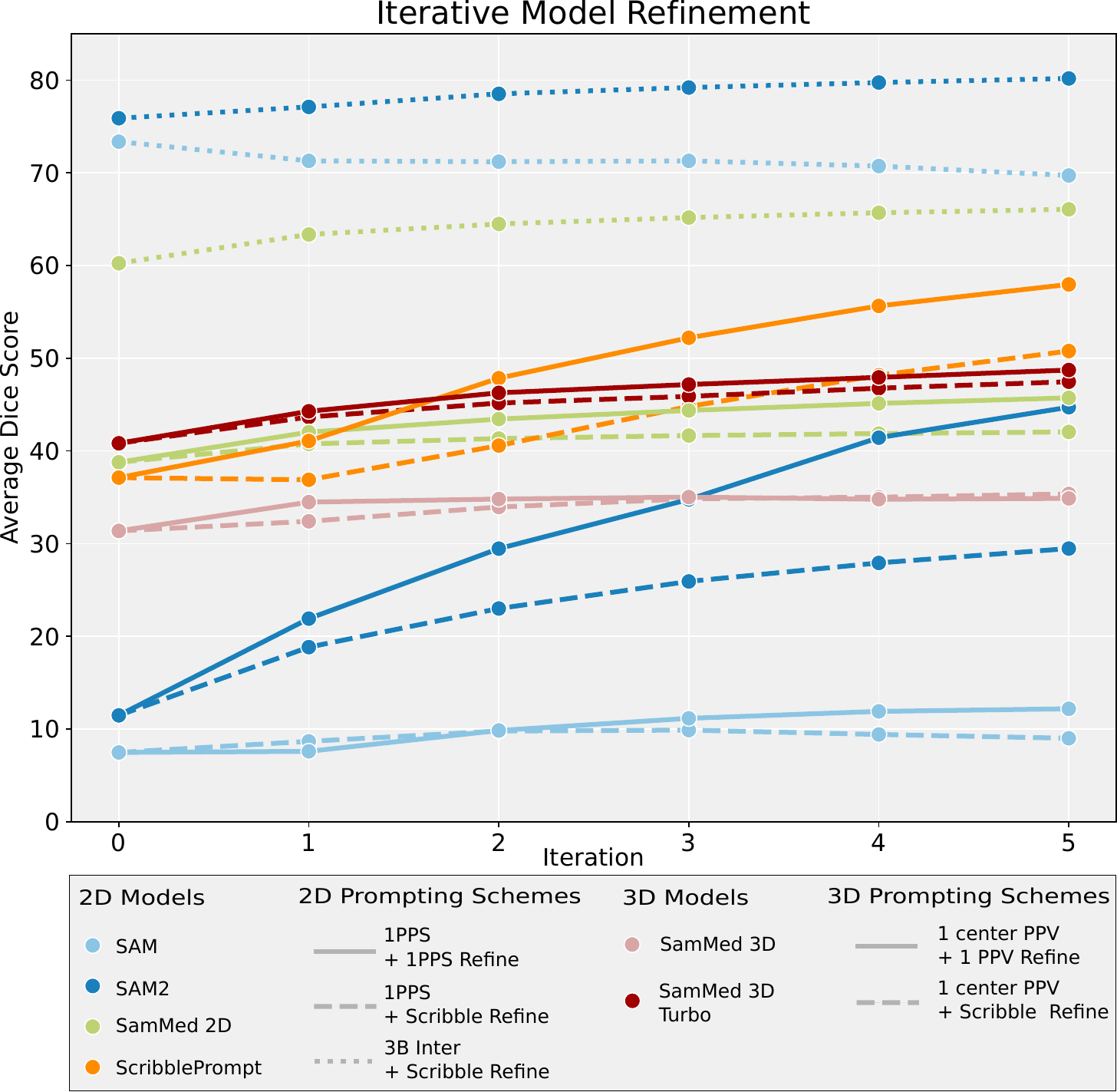}
    \caption{\textbf{Iterative refinement is essential}. All methods improve using the proposed scribble refinement (dashed) or using a random refinement point per volume or slice (solid line). When given a strong initial prediction, as with our proposed 3-box interpolation prompting, 2D models achieve superior performance.}
    \label{fig:refine}
\end{figure}

Finally, we evaluate the performance of the models during iterative refinement. For 2D models, this involves predicting on a slice-by-slice basis. As illustrated in \cref{fig:refine}, adding refinement prompts to each slice results in a substantial performance boost. While the proposed scribble-based refinement consistently improves outcomes, it does not achieve the same level of improvement as adding a prompt to every slice, which is expected since not all imperfect slices are guaranteed to receive new prompts during the scribble refinements. Notably, SAM2 improves more compared to SAM. We observed that for 2D models, it is crucial to provide the initial prompts again for each of the refinement steps. In the appendix in \cref{tab:iterative2d}, we present results without including the previous point in the iterative refinement step, which resulted in a performance decline during refinement. 
As demonstrated in our previous experiments, point prompts do not produce reliable initial segmentations. Although their refinement consistently enhances results, overall performance remains suboptimal. To overcome this, we introduce a new baseline for future interactive segmentation methods: initializing 2D models with three interpolated 2D boxes, followed by Scribble refinement. This approach achieves the best overall performance with Sam2.
\\
For 3D models, iterative refinement also led to consistent performance improvements. Both randomly sampled prompts and those derived from refinement scribbles improved performance similarly. However, they do not reach similar performance as 2D models. 

%% file: sec/4_summary.tex
\section{Discussion and Conclusion}
In this paper, we introduced \itb~ and used it to compare the performance of 2D and 3D interactive segmentation models in 3D medical imaging. We provide a holistic and transparent overview of the current state-of-the-art and highlight key findings that offer practical insights, with additional findings detailed in the appendix \cref{apx:additional_results}:

\begin{enumerate}
    \item \textbf{Realistic 2D prompting can match unrealistic prompting:} Our introduced realistic prompting styles can match unrealistic 2D prompting methods, see \cref{fig:realistic_static}. This unlocks 2D methods for actual 3D medical image workflows without any major performance penalties. 
    \item \textbf{SAM 2 outperforms all medical 2D \& 3D models:} When prompting SAM2 through a realistic prompting scheme such as the Box Interpolation SAM2 is superior to all medical 2D \& 3D models. \textit{Future studies should benchmark against SAM2 with 3-box interpolation and scribble refinement as the new gold standard.}
    \item \textbf{Iterative Refinement is essential:} The ability to iteratively refine segmentations significantly enhances performance, particularly in challenging cases. Models that allow multiple rounds of corrections show better accuracy, making this feature crucial for clinical applications. For example, SegVol reached the highest performance in a static setting for point prompts, however, SAM-Med 3D Turbo can match SegVol given a few interactions.
    \item \textbf{Points fail for difficult and small structures.} Contrary to claims in previous literature \citep{sammed2d,sammd}, point-based methods fail, likely due to previous work training and evaluating their methods on simpler target structures \citep{medsam, sammed2d, sammd}.
    \item \textbf{Bounding boxes outperform points.} Consistent with previous studies \cite{sammd,medsam,segvol,mazurowski2023segment}, we find that Bounding boxes consistently outperform point-based inputs by providing better spatial context, which leads to improved segmentation accuracy, especially for complex structures in radiological images. Point-based prompts lack this context, resulting in poorer performance.
\end{enumerate}

\paragraph{Limitations}
While the \itb~benchmark addresses many issues, our work has limitations. Currently, we only approximate \textit{Human Effort}, providing valuable insights but not fully capturing the complexity of real clinical applications. A crucial next step is a comprehensive study with clinicians to assess the efficacy of different prompting strategies in real-world settings. Beyond segmentation performance, such a study should measure annotation time to evaluate practical feasibility. Ultimately, integrating the best prompting methods into an actual annotation toolkit will be key to improving efficiency in clinical practice.
While the datasets used in our study contain relatively few images, our benchmark includes more datasets than any previous study (\NDatasets~datasets) and covers a wide range of pathologies and anatomical structures in MRI and CT. Moreover, we denote that most larger publicly available datasets have already been used for training, leaving limited options to choose datasets from.
Another limitation is that, given the rapid development of new interactive segmentation methods, very recent approaches are not yet included. However, this also highlights the need for a fair and open benchmark to keep up with ongoing advancements \cite{Marinov}. The published codebase is designed to seamlessly integrate new prompting strategies and models, enabling rapid adaptation and extension.

\paragraph{Conclusion} \itb~ serves as a catalyst for interactive segmentation research in 3D medical imaging by enabling a fair and reproducible comparison between existing methods by its open-source nature. Future work can build upon and extend this framework to investigate crucial questions in interactive segmentation. Through this, \itb~will contribute to improving real-world clinical application of interactive segmentation methods, facilitating a reduction of labor for medical professionals and accelerating clinical research.

%% file: sec/X_suppl.tex
\clearpage
\setcounter{page}{1}
\maketitlesupplementary

\section{Additional Findings}
\label{apx:additional_results}

\begin{figure}*
    \centering
    \includegraphics[width=1\linewidth]{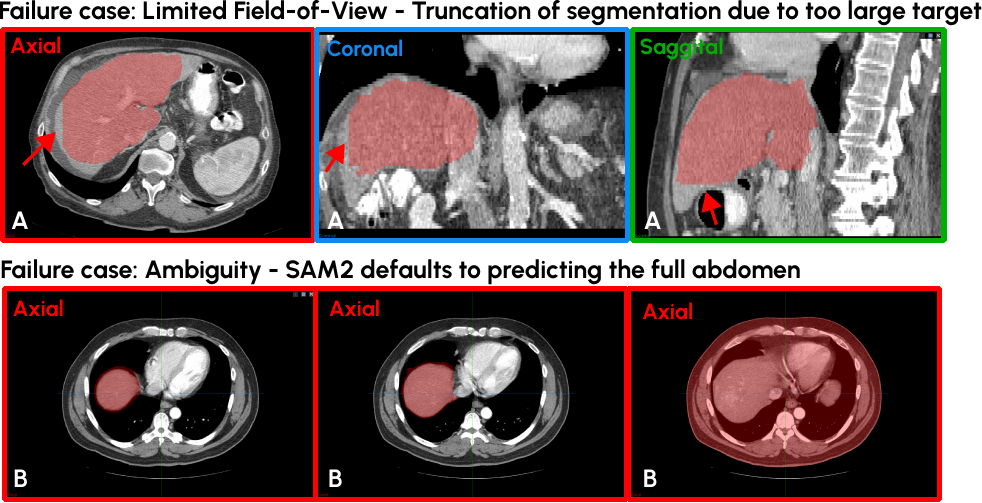}
    \caption{\textbf{Qualitative examples of failure-cases:} A) We highlight issues of SAM-Med3D, which is limited to predict within a small field-of-view. Segmentations are truncated at the edges, due to the organ exceeding the boundaries of the prediction volume. B) 2D segmentation models sometimes default to predicting the entire abdomen, due to the ambiguity of what was queried.}
    \label{fig:qualitative_failure_case}
\end{figure}

\paragraph{Difference between MRI und CT data} \Cref{tab:realisticstatic_modalities} presents the DSC for prompting schemes simulating realistic \textit{Human Effort} without refinement, averaged over MRI datasets (D1-3) and CT datasets (D4-10). While direct comparisons between modalities are not possible due to differing tasks and datasets, notable trends emerge. Some models exhibit consistent performance across modalities, such as SAM2, which shows minimal variation. In contrast, all 3D models display significant performance differences between MRI and CT. This disparity likely reflects the training data distribution; for instance, SegVol was exclusively trained on CT data, leading to subpar performance on MRI datasets. However, on CT-specific tasks, SegVol matches the performance of SAM and SAM2, despite its overall lower average performance across all datasets.

\paragraph{Previous prompts are essential for 2D model refinement.} For 2D models, it is important to reintroduce the initial prompts at each refinement step. In \cref{tab:iterative2d}, we demonstrate that omitting the initial prompt during iterative refinement results in a performance decline. This behavior arises because 2D models often over-segment the target, filling the entire slice foreground. This is visualized in \cref{fig:qualitative_failure_case} B. Without the initial prompt, these models lose crucial target location information, as the initial predicted mask is typically highly inaccurate. Consequently, the refinement process likely generates erroneous additional prompts due to the abundance of false-positive pixels, further degrading performance.

\paragraph{SamMed 3D and SamMed 3D Turbo react less to refinement prompts.} SegVol does not support iterative refinement since it does not accept mask prompts. SamMed 3D and SamMed 3D Turbo are trained exclusively with a single point prompt as input, which we believe makes iterative refinement more challenging. As with the issue mentioned above for 2D models, this limitation can lead to losing track of the target structure. 

\paragraph{Performance comparison between organs and pathological structures}  
\Cref{tab:organvspatho} presents a performance comparison between pathological and organ targets. While a direct comparison is challenging due to variations in tasks and datasets, certain trends can be observed. Notably, 2D models exhibit greater difficulty in segmenting pathologies when prompted with points alone, likely due to the typically higher contrast of organs compared to pathological structures.

\paragraph{Challenges of 3D patch-based prediction for larger structures}  
\Cref{tab:organvspatho} shows the DSC for liver segmentation (D8a), a common task included in the training datasets of all medical models. However, SAM-Med 3D and SAM-Med 3D Turbo encounter difficulties in predicting the entire liver within a single patch. In contrast, SegVol's \textit{zoom-out/zoom-in} inference approach effectively addresses this limitation, achieving a superior DSC of 91\%. Such a failure-case is visualized in \cref{fig:qualitative_failure_case} A.

\paragraph{ScribblePromt performs better with points than with boxes.} We believe this is because, despite being trained on various prompt types, the method is primarily designed for scribble-based interactions, as its name suggests. Since points are more similar to scribbles, they achieve better performance than boxes. We did not include optimized scribble prompts as the initial interaction since other methods do not support them. Notably, ScribblePrompt reacts strongly to multiple points, which may resemble a scribble in interaction style, leading to a significant performance improvement.

\paragraph{SAM reacts less to additional promts than SAM2.} Both for multiple initial point promts, as shown in \cref{tab:unrealistic}, and for iterative refinement, as shown in \cref{tab:iterative2d}, SAM2 improves more with additional prompts. 

\section{Current State of Independent Benchmarkings}
\label{apx:sec:literature_review}
SAM’s introduction spurred numerous efforts to use it for medical applications, leading to a wave of studies evaluating its performance on medical data. Early benchmarks primarily compared SAM with non-interactive models \cite{SAMBenchmarkHe}, \cite{sammd} or outdated interactive methods \cite{mazurowski2023segment}. More recently, an influx of medical-specific SAM adaptations has emerged \cite{medsam,sammed2d,sammed3d,segvol,rokuss2025lesionlocator,ScribblePrompt}. While these studies present comparisons with baseline methods, they tend to focus on evaluation settings where their approach excels. Independent benchmarking remains essential for obtaining an unbiased and comprehensive assessment of relative performance \cite{Marinov}, particularly given the opaque and insufficient evaluation practices observed in some of these works (P3). \\
Currently, most existing benchmarkings focus on analyzing the behavior of only one or two models, mostly SAM and SAM2 \cite{zhang2024unleashingpotentialsam2biomedical, antonov2024rclicksrealisticclicksimulation,review_sam,sammd}, or only evaluating 2D images \cite{IMIS,antonov2024rclicksrealisticclicksimulation,review_sam,sammd}. While the comprehensive 2D medical benchmarking study by Chen et al. \cite{IMIS} incorporates multiple recent methods, it has several limitations. Notably, it evaluates only three external datasets, two of which (SegThor and ISLES2022) were used in the training of both MedSAM and SAM-Med2D. Additionally, the evaluation focuses on larger, homogeneous target structures, lacking smaller, more challenging pathological structures that are difficult to delineate.

\noindent To our knowledge, no existing work has specifically addressed a realistic scenario for 3D interactive segmentation. This highlights a gap in the current benchmarking literature, which has yet to catch up with recent advancements in interactive medical segmentation. Given the rapid development of new 2D and 3D models expected in the coming months and years, this remains a significant challenge. Our independent benchmarking not only updates the field but also introduces an open-source framework that enables future benchmarking efforts to be conducted more efficiently, transparently, and comparably.

\section{Supported prompting schemes}
\label{apx:sec:prompt_scheme_details}

\paragraph{2D initial prompts:}
For the 2D models, every foreground-containing slice needs an initial prompt. \itb~supports the following 2D prompting schemes:
\begin{itemize}
    \item \textit{N} PPS: For each foreground-containing slice, \textit{N} random foreground points are selected.
    \item \textit{N}$\pm$PPS: For each foreground-containing slice, $\lceil N/2 \rceil$ random foreground and $\lfloor N/2 \rfloor$ background points are selected. 
    \item Box PS: For each foreground-containing slice, we select a bounding box around the foreground. 
    \item \textit{N}P Inter: \textit{N} points are used for the proposed point interpolation prompting scheme as explained in \cref{subsec:InitialPrompting}. At least 3 points are needed. 
    \item P Prop: The \textit{center} point of the slice with index $I_{m}$ is selected, and passed to the model along with $min(I)$ and $max(I)$ with notation and method as explained in \cref{subsec:InitialPrompting}.
    \item \textit{N}B Inter: \textit{N} 2D boxes are used for the proposed box interpolation prompting scheme as explained in \cref{subsec:InitialPrompting}. At least 3 boxes are needed. 
    \item B Prop: The 2D box from the slice with index $I_m$ is selected, and passed to the model along with $min(I)$ and $max(I)$ with notation and method as explained in\cref{subsec:InitialPrompting}.
\end{itemize}

\paragraph{3D initial prompts}
For 3D models, the entire volume is predicted in a single step, making the prompting process more accessible for human interaction.
\begin{itemize}
    \item \textit{N} PPV: \textit{N} random foreground points are selected from the volume.
    \item \textit{N} center PPV: We generate a set of slice points using the point interpolation method as in \cref{subsec:InitialPrompting} and then select a subset o $N$ points spaced equally along the axial direction.
    \item 3D Box: A simple 3D bounding box around the foreground is used. 
\end{itemize}

\paragraph{2D iterative refinement}
The refinement experiments always consist of an initial prompt, followed by iterative refinement prompts that are generated as explained in the following. For all refinement steps, a point prompt from a previous step can be included. 
\begin{itemize}
    \item 1PPS Refine: For each slice, a refinement point prompt is selected randomly from the misclassified pixels, and passed either a false positive or false negative as appropriate from the false positive or false negative predicted volume.
    \item Scribble Refine: This prompting scheme follows the proposed scribble refinement scheme for 2D models in \cref{subsec:refinementprompting}. Only slices that receive a new refinement prompt are predicted again.
\end{itemize}

\paragraph{3D iterative refinement}
Again, a point prompt from a previous step can be reused in all subsequent refinement steps.
\begin{itemize}
    \item  1 PPV Refine: For each refinement step, one positive or negative point prompt per volume is selected randomly from the misclassified pixels, and passed either a false positive or false negative as appropriate. 
    \item Scribble Refine: We generate a refinement scribble as we would for 2D iterative refinement, but then randomly select just one refinement point from the scribble. \\
\end{itemize}

All 2D and 3D bounding boxes were generated with perfect alignment around the target object. Since the impact of non-perfect bounding boxes has been extensively studied in previous works \cite{sammd,mazurowski2023segment}, we only explore this additional degree of freedom for the proposed new gold standard Sam2 with 3 Box interpolation in \cref{tab:unrealistic}.

\begin{figure}*
    \centering
    \includegraphics[width=1\linewidth]{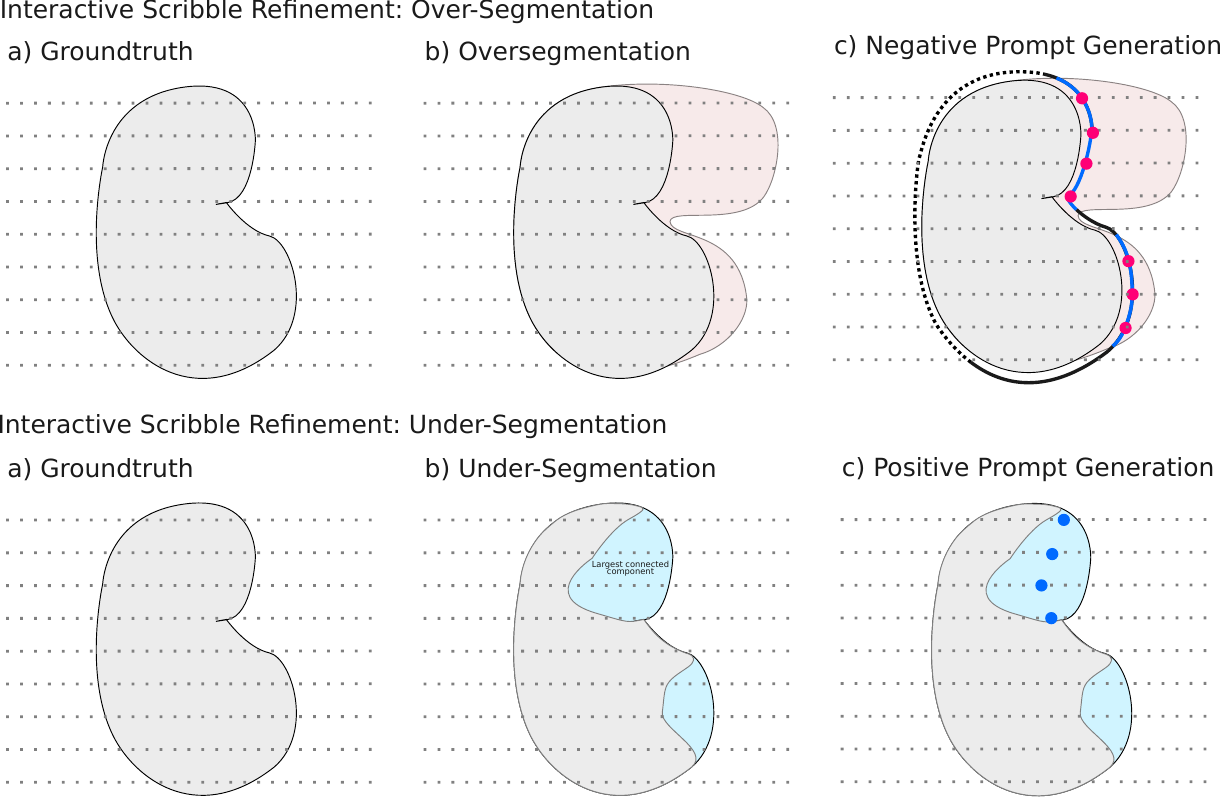}
    \caption{\textbf{Realistic Scribble Refinement.} When the previous segmentation overestimates the target, we identify the non-axial slice with the highest number of false positives and generate a contour around the ground truth object at a 2-pixel distance. A 60\% segment of this contour is then selected, using all false positive points along it as negative refinement prompts for 2D models, while a random FP point of the curve is chosen for 3D models. Conversely, when the segmentation underestimates the target, we perform a connected component analysis on false negative voxels and generate a "scribble" through the center of the largest false negative component. Each axial point along this scribble serves as a new positive prompt for 2D models and one random point of this scribble for 3D models.}
    \label{fig:prompting_refine}
\end{figure}

\section{Model specifications}

\subsection{SAM}
SAM is compatible with multiple image encoders, particularly the ViT family from \cite{transformers}. We used the default and best-performing model with ViT-Huge. To ensure high-quality inputs for the model, we performed slice-wise inference by extracting slices from the inplane-plane axis. Each slice was normalized by first clipping values outside the 0.5th and 99.5th percentile of the volume's intensity distribution and then scaling the values to [0, 255]. The image was repeated three times along the channel axis to produce an RGB-like image. Internally, SAM resizes these slices to 1024 pixels for the longest side with the shorter side being padded to 1024 pixels if needed to maintain square dimensions. Finally, the images are normalized using the model’s pre-stored mean and standard deviation as suggested by the original implementation. Inference was restricted to slices containing foreground. After prediction, the slices were reassembled into a volume, inverse transformed to the original coordinate system, and metrics were computed in the original image space. 

\subsection{SAM2}
SAM2 supports multiple image encoders, specifically the Hiera family of \citet{hiera}. We used the best-performing model, Hiera-L. We clip the intensity values of the volumes based on the 0.5th and 99.5th percentiles, extract each slice along the through-plane, and make the images RGB-like just as with SAM. The images are then rescaled to $1024\times1024$ pixels and again normalized using the mean and standard deviation provided together with the pretrained weights. Aggregation and inverse transformation are then performed similarly to SAM.

\subsection{MedSAM}
To apply the model slice-wise, we slice the input volume as with SAM, and then clip each slice based on their 0.5th and 99.5th percentile values. The images are then made RGB-like by repeating thrice along a new channel-dimension, rescaled to $1024\times1024$ pixels and then normalised to [0, 1]. Aggregation and inverse transformation are performed similarly as with SAM.

\subsection{SAM-Med2D}
To apply the model slice-by-slice, we slice the input volume as with SAM, and then clip each slice based on their 0.5th and 99.5th percentile values same as with MedSAM. The slices are then made RGB-like and converted to a [0, 255] scale as in SAM's preprocessing. The slices are then standardized using a mean and standard deviation provided along with the model and resized to $256\times256$ pixels. Aggregation and inverse transformation are performed similarly as with SAM.

\subsection{SAM-Med3D}
The model is 3D so no slicing is needed. The volume is respaced to $1.5 \times 1.5 \times 1.5$ mm and then clipped based on its 0.5th and 99.5th percentiles. SAM-Med 3D performs inference on a 128x128x128 crop. The crop is centered around our point prompt if there is only one point prompt passed, and around the centroid of our prompts if multiple points are passed simultaneously. For subsequent refinement steps, the crop remains unchanged. The predicted crop is inserted back in its correct position within the wider coordinate system and then respaced back to the original spacing so that evaluation takes place in the corresponding native image space.

\subsection{SAM-Med3D Turbo}
SAM-Med3D Turbo is an updated checkpoint for SAM-Med-3D and so we perform the same pre- and postprocessing.

\subsection{SegVol}
Intensity values are clipped by its 0.5th and 99.5th percentiles. The mean and standard deviation of the foreground voxels are used for zscore normalization. The values are then rescaled to a [0,1]. Finally, the volume is cropped to its foreground. A first 'zoom-out' inference is performed on this image, followed by a 'zoom-in'  sliding window inference. The predicted volume is then transformed back to the original space and compared with the unprocessed ground truth to calculate metrics.

\subsection{ScribblePromt}
Each image slice is normalized between 0 and 1 and reshaped to a fixed 128x128 image size. Inference was restricted to slices containing foreground. The predicted slices were inverse transformed to the original coordinate system, and metrics were computed in the original image space. Notably, we apply ScribblePrompt only on axial slices, but it was originally trained on all views.

\begin{table*}[!htbp]
    \centering
    
    \include{results/unrealistic_static.tex}

    \label{tab:unrealistic}
\end{table*}
\FloatBarrier
\clearpage

\begin{table*}[!htbp]
    \centering
    \include{results/realistic_static_2D.tex}
    \label{tab:realistic_effort_2d}
\end{table*}
\FloatBarrier

\begin{table*}[!htbp]
    \centering
    \include{results/realistic_static_3D.tex}
    \label{tab:realistic_effort_3d}
\end{table*}
\FloatBarrier
\clearpage
\begin{table*}[!htbp]
    \centering
    \include{results/interactive_2D}

    \label{tab:iterative2d}
\end{table*}
\clearpage
\begin{table*}[!htbp]
    \centering
    \include{results/interactive_3D}
    \label{tab:iterative3d}
\end{table*}

\begin{table*}[!htbp]
    \centering
    
    \include{results/organ_vs_patho}
    \label{tab:organvspatho}
\end{table*}
\begin{table*}[!htbp]
    \centering

\include{results/realistic_static_modalities}
    \label{tab:realisticstatic_modalities}
\end{table*}
\begin{table*}[!htbp]
    \centering
    
    \include{results/uncertainty_boxes}
    \label{tab:uncertainty_boxes}
\end{table*}

%% file: results/unrealistic_static.tex
\resizebox{\linewidth}{!}{
\centering
\begin{tabular}{llllrrrrrrrrrr}
\toprule
Prompter & Model & Interactions & D1 & D2 & D3 & D4 & D5 & D6 & D7 & D8 & D9 & D10 & Average \\
\midrule
\rowcolor[gray]{1}
SAM & 1PPS & 1X & NaN & 3.42 & 3.06 & 4.21 & 1.43 & 0.92 & NaN & 15.4 & 28.52 & 4.89 & 7.73 \\
\rowcolor[gray]{0.95}
SAM2 & 1PPS & 1X & 3.36 & 4.99 & 4.41 & 5.91 & 3.77 & 1.62 & 15.64 & 22.36 & 45.86 & 6.79 & 11.47 \\
\rowcolor[gray]{0.9}
SamMed 2D & 1PPS & 1X & 18.28 & 33.98 & 44.01 & 41.44 & 35.97 & 18.79 & 59.09 & 49.33 & 62.66 & 24.28 & 38.78 \\
\rowcolor[gray]{0.85}
ScribblePrompt & 1PPS & 1X & 25.29 & 32.9 & 47.38 & 49.09 & 26.43 & 4.37 & 49.73 & 45.4 & 55.47 & 35.12 & 37.12 \\
\rowcolor[gray]{1}
SAM & 2PPS & 2X & 1.98 & 3.59 & 3.16 & 6.22 & 2.18 & 0.99 & 12.29 & 17.34 & 28.95 & 5.43 & 8.21 \\
\rowcolor[gray]{0.95}
SAM2 & 2PPS & 2X & 4.4 & 5.83 & 5.8 & 6.67 & 4.41 & 1.62 & 17.82 & 22.35 & 48.72 & 7.63 & 12.53 \\
\rowcolor[gray]{0.9}
SamMed 2D & 2PPS & 2X & 17.99 & 37.44 & 44.99 & 44.82 & 39.0 & 20.38 & 68.51 & 54.9 & 67.2 & 27.13 & 42.24 \\
\rowcolor[gray]{0.85}
ScribblePrompt & 2PPS & 2X & 37.15 & 43.31 & 60.66 & 67.58 & 39.55 & 14.9 & 59.35 & 51.42 & 65.31 & 43.83 & 48.31 \\
\rowcolor[gray]{1}
SAM & 2$\pm$PPS & 2X & 3.47 & 3.91 & 3.8 & 7.47 & 2.19 & 1.31 & 15.5 & 19.88 & 38.33 & 6.93 & 10.28 \\
\rowcolor[gray]{0.95}
SAM2 & 2$\pm$PPS & 2X & 11.24 & 10.86 & 12.73 & 11.0 & 6.42 & 2.87 & 38.04 & 29.13 & 54.47 & 11.18 & 18.79 \\
\rowcolor[gray]{0.9}
SamMed 2D & 2$\pm$PPS & 2X & 19.62 & 36.56 & 47.23 & 45.41 & 37.39 & 21.44 & 61.12 & 50.44 & 64.5 & 25.06 & 40.88 \\
\rowcolor[gray]{0.85}
ScribblePrompt & 2$\pm$PPS & 2X & 34.08 & 37.68 & 50.98 & 64.55 & 35.92 & 24.43 & 54.23 & 47.68 & 57.72 & 35.07 & 44.23 \\
\rowcolor[gray]{1}
SAM & 3PPS & 3X & 2.07 & 4.0 & 3.38 & 7.22 & 2.92 & 1.19 & 14.27 & 19.36 & 31.1 & 5.84 & 9.13 \\
\rowcolor[gray]{0.95}
SAM2 & 3PPS & 3X & 6.19 & 8.26 & 9.35 & 9.29 & 5.72 & 1.99 & 28.66 & 24.23 & 53.18 & 8.32 & 15.52 \\
\rowcolor[gray]{0.9}
SamMed 2D & 3PPS & 3X & 17.61 & 38.93 & 45.34 & 44.74 & 38.86 & 21.14 & 73.81 & 58.24 & 69.45 & 27.9 & 43.6 \\
\rowcolor[gray]{0.85}
ScribblePrompt & 3PPS & 3X & 43.23 & 50.1 & 67.63 & 74.49 & 46.89 & 32.53 & 68.6 & 59.62 & 70.65 & 48.07 & 56.18 \\
\rowcolor[gray]{1}
SAM & 3$\pm$PPS & 3X & 3.82 & 4.33 & 4.12 & 9.38 & 2.87 & 1.42 & 17.77 & 21.84 & 37.91 & 7.13 & 11.06 \\
\rowcolor[gray]{0.95}
SAM2 & 3$\pm$PPS & 3X & 13.75 & 12.49 & 15.33 & 14.66 & 8.0 & 3.24 & 44.98 & 30.34 & 57.62 & 11.75 & 21.22 \\
\rowcolor[gray]{0.9}
SamMed 2D & 3$\pm$PPS & 3X & 19.59 & 39.52 & 47.98 & 47.76 & 40.27 & 22.82 & 71.01 & 56.28 & 68.38 & 27.77 & 44.14 \\
\rowcolor[gray]{0.85}
ScribblePrompt & 3$\pm$PPS & 3X & 43.19 & 47.89 & 62.72 & 73.52 & 48.57 & 43.65 & 65.73 & 58.14 & 66.71 & 44.67 & 55.48 \\
\rowcolor[gray]{1}
SAM & 5PPS & 5X & 4.19 & 5.56 & 5.94 & 11.05 & 4.12 & 1.52 & 22.86 & 23.88 & 40.75 & 7.88 & 12.77 \\
\rowcolor[gray]{0.95}
SAM2 & 5PPS & 5X & 8.68 & 12.59 & 15.58 & 14.89 & 7.32 & 2.68 & 48.72 & 27.08 & 55.08 & 9.76 & 20.24 \\
\rowcolor[gray]{0.9}
SamMed 2D & 5PPS & 5X & 17.06 & 39.74 & 45.49 & 42.65 & 37.08 & 21.57 & 78.73 & 62.38 & 71.48 & 27.74 & 44.39 \\
\rowcolor[gray]{0.85}
ScribblePrompt & 5PPS & 5X & 48.0 & 57.41 & 72.75 & 77.77 & 54.06 & 51.98 & 76.77 & 69.31 & 75.79 & 52.39 & 63.62 \\
\rowcolor[gray]{1}
SAM & 5$\pm$PPS & 5X & 5.17 & 4.43 & 5.74 & 12.05 & 2.87 & 1.78 & 24.41 & 27.98 & 42.64 & 8.51 & 13.56 \\
\rowcolor[gray]{0.95}
SAM2 & 5$\pm$PPS & 5X & 20.17 & 19.59 & 24.07 & 25.43 & 12.91 & 6.17 & 62.15 & 38.99 & 64.65 & 17.33 & 29.15 \\
\rowcolor[gray]{0.9}
SamMed 2D & 5$\pm$PPS & 5X & 20.05 & 41.94 & 49.86 & 49.43 & 40.72 & 24.75 & 76.75 & 60.72 & 70.94 & 28.65 & 46.38 \\
\rowcolor[gray]{0.85}
ScribblePrompt & 5$\pm$PPS & 5X & 50.01 & 56.37 & 68.66 & 77.95 & 58.49 & 58.6 & 69.59 & 64.63 & 72.25 & 49.98 & 62.65 \\
\rowcolor[gray]{1}
SAM & 10PPS & 10X & 5.5 & 6.64 & 8.14 & 15.7 & 4.88 & 2.14 & 31.65 & 28.55 & 47.06 & 16.5 & 16.68 \\
\rowcolor[gray]{0.95}
SAM2 & 10PPS & 10X & 6.67 & 13.62 & 14.5 & 16.1 & 8.13 & 3.45 & 55.1 & 31.33 & 54.52 & 9.38 & 21.28 \\
\rowcolor[gray]{0.9}
SamMed 2D & 10PPS & 10X & 15.94 & 38.7 & 44.96 & 38.15 & 32.23 & 22.22 & 82.57 & 67.07 & 71.92 & 25.7 & 43.95 \\
\rowcolor[gray]{0.85}
ScribblePrompt & 10PPS & 10X & 51.46 & 65.06 & 76.14 & 79.34 & 60.15 & 60.46 & 77.17 & 73.45 & 80.27 & 56.28 & 67.98 \\
\rowcolor[gray]{1}
SAM & 10$\pm$PPS & 10X & 9.75 & 8.0 & 10.6 & 23.07 & 9.61 & 2.02 & 28.43 & 35.11 & 45.62 & 22.77 & 19.5 \\
\rowcolor[gray]{0.95}
SAM2 & 10$\pm$PPS & 10X & 28.7 & 29.26 & 36.21 & 40.99 & 24.98 & 18.53 & 71.71 & 46.53 & 72.77 & 24.55 & 39.42 \\
\rowcolor[gray]{0.9}
SamMed 2D & 10$\pm$PPS & 10X & 21.05 & 43.97 & 51.95 & 50.31 & 38.98 & 27.87 & 81.8 & 66.4 & 72.97 & 28.31 & 48.36 \\
\rowcolor[gray]{0.85}
ScribblePrompt & 10$\pm$PPS & 10X & 55.75 & 66.81 & 76.11 & 82.16 & 68.4 & 69.71 & 72.94 & 68.89 & 78.28 & 57.27 & 69.63 \\
\rowcolor[gray]{1}
SAM & Box PS & 2X & 75.93 & 65.2 & 82.79 & 83.68 & 76.67 & 80.97 & 89.06 & 84.17 & 90.23 & 71.01 & 79.97 \\
\rowcolor[gray]{0.95}
SAM2 & Box PS & 2X & 78.0 & 72.01 & 86.57 & 88.5 & 81.58 & 85.21 & 91.15 & 89.01 & 89.0 & 75.28 & 83.63 \\
\rowcolor[gray]{0.9}
SamMed 2D & Box PS & 2X & 40.41 & 56.79 & 79.73 & 72.84 & 58.12 & 59.11 & 88.97 & 82.2 & 82.71 & 48.01 & 66.89 \\
\rowcolor[gray]{0.85}
ScribblePrompt & Box PS & 2X & 14.68 & 29.2 & 53.97 & 65.9 & 24.61 & 12.4 & 25.53 & 23.03 & 65.46 & 10.39 & 32.52 \\
\rowcolor[gray]{0.8}
MedSam & Box PS & 2X & 58.22 & 69.98 & 72.99 & 76.98 & 69.9 & 68.18 & 92.87 & 89.74 & 68.1 & 55.95 & 72.29 \\
\bottomrule
\end{tabular}}
\caption{\textbf{Experimental results simulating unrealistic effort of a clinician}. This involves prompting each slice of a 3D volume. 'PPS' and 'BPS' represent points per slice or box per slice, respectively. 'X' implies that each interaction is replicated for every slice, multiplying the clinician's effort across the entire volume. }

%% file: results/realistic_static_2D.tex
\centering
\resizebox{\linewidth}{!}{
\begin{tabular}{llllrrrrrrrrrr}
\toprule
Prompter & Model & Interactions & D1 & D2 & D3 & D4 & D5 & D6 & D7 & D8 & D9 & D10 & Average \\
\midrule
\rowcolor[gray]{1}
SAM & 3P Inter & 3 & 2.06 & 3.65 & 3.06 & 4.81 & 1.41 & 0.91 & 11.05 & 15.45 & 14.81 & 6.63 & 6.38 \\
\rowcolor[gray]{0.95}
SAM2 & 3P Inter & 3 & 3.76 & 5.39 & 4.5 & 6.52 & 3.74 & 1.55 & 15.6 & 22.57 & 23.96 & 9.22 & 9.68 \\
\rowcolor[gray]{0.9}
SamMed 2D & 3P Inter & 3 & 19.13 & 34.07 & 44.69 & 47.93 & 42.86 & 20.63 & NaN & 49.54 & 55.82 & 27.38 & 38.01 \\
\rowcolor[gray]{0.85}
ScribblePrompt & 3P Inter & 3 & 27.69 & 34.7 & 56.69 & 59.87 & 28.6 & 4.58 & 50.46 & 45.5 & 52.22 & 41.4 & 40.17 \\
\rowcolor[gray]{1}
SAM & 5P Inter & 5 & 2.05 & 3.57 & 3.07 & 4.84 & 1.41 & 0.91 & 11.04 & 15.43 & 20.65 & 6.7 & 6.97 \\
\rowcolor[gray]{0.95}
SAM2 & 5P Inter & 5 & 3.77 & 5.37 & 4.52 & 6.83 & 3.76 & 1.55 & 15.61 & 22.59 & 32.41 & 9.12 & 10.55 \\
\rowcolor[gray]{0.9}
SamMed 2D & 5P Inter & 5 & 19.27 & 34.86 & 44.91 & 48.21 & 43.35 & 20.77 & 64.46 & 49.73 & 60.7 & 27.6 & 41.39 \\
\rowcolor[gray]{0.85}
ScribblePrompt & 5P Inter & 5 & 27.93 & 35.68 & 57.27 & 59.45 & 28.82 & 4.58 & 50.77 & 45.39 & 57.67 & 41.49 & 40.9 \\
\rowcolor[gray]{1}
SAM & 10P Inter & 10 & 2.05 & 3.6 & 3.06 & 4.8 & 1.41 & 0.92 & 11.05 & 15.46 & 25.25 & 6.48 & 7.41 \\
\rowcolor[gray]{0.95}
SAM2 & 10P Inter & 10 & 3.77 & 5.38 & 4.51 & 6.72 & 3.74 & 1.55 & 15.71 & 22.5 & 40.64 & 8.7 & 11.32 \\
\rowcolor[gray]{0.9}
SamMed 2D & 10P Inter & 10 & 19.33 & 35.48 & 44.75 & 48.18 & 43.56 & 20.73 & 64.51 & 48.64 & 62.7 & 27.67 & 41.56 \\
\rowcolor[gray]{0.85}
ScribblePrompt & 10P Inter & 10 & 28.01 & 36.45 & 57.6 & 59.56 & 28.84 & 4.61 & 50.95 & 44.81 & 59.56 & 41.52 & 41.19 \\
\rowcolor[gray]{1}
SAM & 5P Prop & 7 & 3.91 & 3.93 & 3.2 & 4.17 & 1.82 & 0.91 & 11.55 & 14.37 & 12.74 & 6.31 & 6.29 \\
\rowcolor[gray]{0.95}
SAM2 & 5P Prop & 7 & 7.82 & 6.37 & 4.66 & 5.57 & 4.44 & 1.63 & 14.89 & 19.61 & 17.25 & 7.93 & 9.02 \\
\rowcolor[gray]{0.9}
SamMed 2D & 5P Prop & 7 & 15.5 & 26.04 & 30.54 & 30.25 & 28.69 & 17.98 & 58.1 & 47.94 & 42.37 & 16.95 & 31.44 \\
\rowcolor[gray]{0.85}
ScribblePrompt & 5P Prop & 7 & 41.08 & 31.39 & 42.63 & 38.61 & 24.88 & 4.85 & 43.78 & 33.65 & 32.1 & 35.27 & 32.82 \\
\rowcolor[gray]{1}
SAM & B Prop & 4 & 4.77 & 37.71 & 53.06 & 65.66 & 46.49 & 43.74 & 69.39 & 50.39 & 76.49 & 40.41 & 48.81 \\
\rowcolor[gray]{0.95}
SAM2 & B Prop & 4 & 4.76 & 42.35 & 60.54 & 67.65 & 50.53 & 49.55 & 76.06 & 67.11 & 74.4 & 39.23 & 53.22 \\
\rowcolor[gray]{0.9}
SamMed 2D & B Prop & 4 & 5.0 & 35.97 & 60.33 & 52.7 & 35.09 & 37.46 & 74.3 & 69.5 & 65.06 & 26.43 & 46.19 \\
\rowcolor[gray]{0.85}
ScribblePrompt & B Prop & 4 & 3.43 & 12.27 & 21.49 & 46.45 & 12.08 & 5.16 & 7.69 & 3.24 & 13.99 & 3.79 & 12.96 \\
\rowcolor[gray]{0.8}
MedSam & B Prop & 4 & 4.8 & 39.38 & 38.71 & 45.48 & 45.54 & 42.95 & 57.25 & 55.32 & 15.47 & 29.11 & 37.4 \\
\rowcolor[gray]{1}
SAM & from 3D Box & 3 & 69.67 & 47.56 & 60.62 & 67.56 & 59.08 & 63.77 & 76.43 & 57.28 & 50.0 & 54.28 & 60.63 \\
\rowcolor[gray]{0.95}
SAM2 & from 3D Box & 3 & 71.05 & 51.5 & 62.5 & 66.23 & 62.89 & 66.93 & 76.65 & 60.62 & 41.27 & 56.31 & 61.6 \\
\rowcolor[gray]{0.9}
SamMed 2D & from 3D Box & 3 & 37.7 & 46.19 & 65.53 & 58.18 & 48.06 & 53.71 & 75.76 & 64.52 & 47.57 & 39.05 & 53.63 \\
\rowcolor[gray]{0.85}
ScribblePrompt & from 3D Box & 3 & 15.45 & 30.41 & 56.52 & 65.43 & 33.7 & 15.38 & 36.47 & 34.42 & 56.19 & 12.61 & 35.66 \\
\rowcolor[gray]{0.8}
MedSam & from 3D Box & 3 & 54.94 & 59.43 & 68.82 & 71.48 & 68.14 & 67.02 & 85.53 & 73.18 & 31.32 & 48.94 & 62.88 \\
\rowcolor[gray]{1}
SAM & 3B Inter & 6 & 75.41 & 60.53 & 80.06 & 81.65 & 75.75 & 77.85 & 86.54 & 76.3 & 62.48 & 67.48 & 74.41 \\
\rowcolor[gray]{0.95}
SAM2 & 3B Inter & 6 & 77.29 & 65.06 & 80.2 & 82.63 & 78.6 & 80.43 & 86.29 & 78.43 & 58.8 & 71.41 & 75.91 \\
\rowcolor[gray]{0.9}
SamMed 2D & 3B Inter & 6 & 39.98 & 52.3 & 75.68 & 68.33 & 56.34 & 56.67 & 80.76 & 69.65 & 58.4 & 44.42 & 60.25 \\
\rowcolor[gray]{0.85}
ScribblePrompt & 3B Inter & 6 & 14.16 & 24.84 & 44.69 & 54.56 & 18.72 & 9.78 & 17.79 & 9.25 & 32.14 & 7.09 & 23.3 \\
\rowcolor[gray]{0.8}
MedSam & 3B Inter & 6 & 57.46 & 60.45 & 61.44 & 65.3 & 64.57 & 61.43 & 72.82 & 67.6 & 38.8 & 49.05 & 59.89 \\
\rowcolor[gray]{1}
SAM & 5B Inter & 10 & 75.8 & 63.21 & 82.1 & 83.59 & 76.32 & 80.07 & 89.29 & 83.54 & 86.57 & 70.25 & 79.08 \\
\rowcolor[gray]{0.95}
SAM2 & 5B Inter & 10 & 77.83 & 69.05 & 84.69 & 87.5 & 80.77 & 83.83 & 90.81 & 87.13 & 82.67 & 74.55 & 81.88 \\
\rowcolor[gray]{0.9}
SamMed 2D & 5B Inter & 10 & 40.34 & 55.13 & 78.8 & 72.31 & 57.63 & 58.42 & 87.67 & 80.17 & 79.12 & 47.26 & 65.68 \\
\rowcolor[gray]{0.85}
ScribblePrompt & 5B Inter & 10 & 14.58 & 27.84 & 51.5 & 63.74 & 23.54 & 11.88 & 23.3 & 16.74 & 54.92 & 9.48 & 29.75 \\
\rowcolor[gray]{0.8}
MedSam & 5B Inter & 10 & 58.09 & 66.79 & 69.7 & 74.97 & 68.81 & 66.75 & 89.11 & 84.64 & 59.6 & 54.28 & 69.27 \\
\rowcolor[gray]{1}
SAM & 10B Inter & 20 & 75.93 & 64.49 & 82.38 & 83.54 & 76.57 & 80.61 & 89.08 & 84.21 & 89.99 & 70.84 & 79.76 \\
\rowcolor[gray]{0.95}
SAM2 & 10B Inter & 20 & 77.99 & 70.88 & 85.95 & 88.26 & 81.44 & 84.8 & 91.02 & 88.61 & 88.28 & 75.08 & 83.23 \\
\rowcolor[gray]{0.9}
SamMed 2D & 10B Inter & 20 & 40.41 & 56.3 & 79.4 & 72.63 & 58.04 & 58.99 & 88.79 & 81.88 & 82.37 & 47.84 & 66.67 \\
\rowcolor[gray]{0.85}
ScribblePrompt & 10B Inter & 20 & 14.68 & 28.75 & 53.45 & 65.72 & 24.56 & 12.33 & 25.12 & 21.38 & 63.31 & 10.26 & 31.96 \\
\rowcolor[gray]{0.8}
MedSam & 10B Inter & 20 & 58.22 & 68.9 & 72.2 & 76.64 & 69.77 & 67.85 & 92.23 & 88.6 & 66.52 & 55.74 & 71.67 \\
\bottomrule
\end{tabular}}
\caption{\textbf{Experimental results simulating a realistic clinician's effort.} 'PPS' and 'PPV' represent points per slice or volume, respectively. 'B Prop' and 'P Prop' denote the introduced box and point propagation schemes, while 'B Inter' and 'P Inter' refer to the introduced box and point interpolation methods.}
\label{tab:realisticstatic}

%% file: results/realistic_static_3D.tex
\centering
\resizebox{\linewidth}{!}{
\centering
\begin{tabular}{llllrrrrrrrrrr}
\toprule
Prompter & Model & Interactions & D1 & D2 & D3 & D4 & D5 & D6 & D7 & D8 & D9 & D10 & Average \\
\midrule
\rowcolor[gray]{1}
SegVol & 1 center PPV & 1 & 9.96 & 24.91 & 38.49 & 31.36 & 3.17 & 33.92 & 71.01 & 50.67 & 28.21 & 30.73 & 32.24 \\
\rowcolor[gray]{0.95}
SamMed 3D Turbo & 1 center PPV & 1 & 5.18 & 27.34 & 46.07 & 34.33 & 15.91 & 46.38 & 82.95 & 59.5 & 63.75 & 26.98 & 40.84 \\
\rowcolor[gray]{0.9}
SamMed 3D & 1 center PPV & 1 & 2.07 & 12.15 & 24.06 & 27.16 & 15.11 & 19.64 & 72.66 & 53.24 & 50.38 & 26.51 & 30.3 \\
\rowcolor[gray]{1}
SegVol & 1PPV & 1 & 8.84 & 21.28 & 31.49 & 25.49 & 2.45 & 32.66 & 61.77 & 52.46 & 29.09 & 25.85 & 29.14 \\
\rowcolor[gray]{0.95}
SamMed 3D Turbo & 1PPV & 1 & 5.03 & 18.34 & 30.08 & 18.38 & 10.75 & 34.93 & 28.51 & 19.26 & 39.26 & 17.06 & 22.16 \\
\rowcolor[gray]{0.9}
SamMed 3D & 1PPV & 1 & 1.92 & 10.95 & 21.19 & 29.12 & 13.28 & 16.66 & 56.06 & 48.55 & 42.81 & 23.63 & 26.42 \\
\rowcolor[gray]{1}
SegVol & 2 center PPV & 2 & 11.2 & 31.31 & 47.51 & 58.45 & 11.57 & 52.36 & 75.08 & 52.66 & 33.4 & 32.45 & 40.6 \\
\rowcolor[gray]{1}
SegVol & 2PPV & 2 & 22.53 & 32.08 & 44.9 & 51.82 & 9.01 & 53.99 & 70.19 & 57.02 & 36.2 & 33.26 & 41.1 \\
\rowcolor[gray]{1}
SegVol & 3 center PPV & 3 & 11.52 & 31.1 & 50.08 & 57.51 & 18.76 & 53.46 & 73.04 & 58.22 & 45.52 & 33.95 & 43.32 \\
\rowcolor[gray]{1}
SegVol & 3PPV & 3 & 25.46 & 37.33 & 51.77 & 62.41 & 16.79 & 59.36 & 77.44 & 62.83 & 47.21 & 36.57 & 47.72 \\
\rowcolor[gray]{1}
SegVol & 5 center PPV & 5 & 11.7 & 31.17 & 49.08 & 52.47 & 25.4 & 52.85 & 61.6 & 45.66 & 44.16 & 33.48 & 40.76 \\
\rowcolor[gray]{1}
SegVol & 5PPV & 5 & 26.07 & 41.98 & 61.5 & 65.13 & 27.73 & 62.78 & 82.23 & 71.76 & 59.96 & 40.47 & 53.96 \\
\rowcolor[gray]{1}
SegVol & 10 center PPV & 10 & 11.69 & 30.22 & 45.24 & 47.32 & 26.43 & 51.68 & 42.04 & 26.69 & 41.88 & 32.96 & 35.61 \\
\rowcolor[gray]{1}
SegVol & 10PPV & 10 & 26.09 & 46.51 & 68.3 & 66.98 & 38.92 & 64.67 & 84.18 & 76.73 & 70.49 & 44.74 & 58.76 \\
\rowcolor[gray]{1}
SegVol & 3D Box & 3 & 0.55 & 37.17 & 68.11 & 69.72 & 63.21 & 50.13 & 89.95 & 79.98 & 72.13 & 49.45 & 58.04 \\
\bottomrule
\end{tabular}}
\caption{Experimental results simulating a realistic clinician's effort. 'center PPV' refers to points sampled at equal axial distances along the five-point interpolation prompter line, which approximates the centerline of the target volume. Sampling from the center performs better than sampling a random point only for the first point. SegVol trained on CT images from the D2 HanSeg dataset, while we tested on MRI images from the same patient cohort.}
\label{tab:realisticstatic}

%% file: results/interactive_2D.tex
    \centering
\resizebox{!}{0.4\textheight}{
\begin{tabular}{llllrrrrrrrrrrr}
\toprule
Model & Promter & Interactions & Iteration &D1 & D2 & D3 & D4 & D5 & D6 & D7 & D8 & D9 & D10 & Average \\
\midrule
\rowcolor[gray]{1}
SAM & 1PPS + 1PPS Refine & 1X/1X & 0 & 1.89 & 3.36 & 3.06 & 4.21 & 1.43 & 0.92 & 11.06 & 15.4 & 28.52 & 4.89 & 7.47 \\
\rowcolor[gray]{1}
SAM & 1PPS + 1PPS Refine & 1X/1X & 1 & 0.1 & 1.11 & 0.17 & 1.07 & 0.01 & 0.02 & 0.43 & 0.71 & 5.57 & 0.75 & 0.99 \\
\rowcolor[gray]{1}
SAM & 1PPS + 1PPS Refine & 1X/1X & 2 & 0.45 & 2.0 & 2.08 & 2.31 & 1.01 & 0.63 & 6.64 & 11.62 & 13.0 & 3.26 & 4.3 \\
\rowcolor[gray]{1}
SAM & 1PPS + 1PPS Refine & 1X/1X & 3 & 0.63 & 2.8 & 1.13 & 2.46 & 0.88 & 0.42 & 4.93 & 6.76 & 9.58 & 2.18 & 3.18 \\
\rowcolor[gray]{1}
SAM & 1PPS + 1PPS Refine & 1X/1X & 4 & 0.43 & 2.08 & 1.95 & 1.76 & 0.67 & 0.6 & 6.3 & 11.17 & 12.4 & 3.18 & 4.05 \\
\rowcolor[gray]{1}
SAM & 1PPS + 1PPS Refine & 1X/1X & 5 & 0.4 & 2.21 & 1.47 & 1.81 & 0.65 & 0.53 & 5.08 & 7.84 & 8.52 & 2.31 & 3.08 \\
\rowcolor[gray]{0.95}
SAM2 & 1PPS + 1PPS Refine & 1X/1X & 0 & 3.36 & 4.99 & 4.41 & 5.91 & 3.77 & 1.62 & 15.64 & 22.36 & 45.86 & 6.79 & 11.47 \\
\rowcolor[gray]{0.95}
SAM2 & 1PPS + 1PPS Refine & 1X/1X & 1 & 3.32 & 5.79 & 6.72 & 6.83 & 4.95 & 1.99 & 23.76 & 28.73 & 47.73 & 6.98 & 13.68 \\
\rowcolor[gray]{0.95}
SAM2 & 1PPS + 1PPS Refine & 1X/1X & 2 & 2.41 & 6.3 & 7.63 & 8.2 & 5.23 & 1.99 & 25.8 & 29.07 & 48.78 & 7.36 & 14.28 \\
\rowcolor[gray]{0.95}
SAM2 & 1PPS + 1PPS Refine & 1X/1X & 3 & 2.4 & 7.11 & 8.9 & 8.39 & 5.35 & 1.98 & 28.33 & 30.34 & 50.66 & 7.59 & 15.1 \\
\rowcolor[gray]{0.95}
SAM2 & 1PPS + 1PPS Refine & 1X/1X & 4 & 2.52 & 7.15 & 9.36 & 8.15 & 5.45 & 1.85 & 29.92 & 30.69 & 51.29 & 7.52 & 15.39 \\
\rowcolor[gray]{0.95}
SAM2 & 1PPS + 1PPS Refine & 1X/1X & 5 & 3.51 & 7.43 & 9.39 & 7.78 & 5.89 & 1.91 & 30.41 & 31.13 & 52.16 & 7.62 & 15.72 \\
\rowcolor[gray]{0.9}
SamMed 2D & 1PPS + 1PPS Refine & 1X/1X & 0 & 18.28 & 33.98 & 44.01 & 41.44 & 35.97 & 18.79 & 59.09 & 49.33 & 62.66 & 24.28 & 38.78 \\
\rowcolor[gray]{0.9}
SamMed 2D & 1PPS + 1PPS Refine & 1X/1X & 1 & 17.56 & 36.81 & 44.97 & 45.81 & 36.19 & 20.8 & 69.07 & 56.35 & 67.4 & 23.5 & 41.85 \\
\rowcolor[gray]{0.9}
SamMed 2D & 1PPS + 1PPS Refine & 1X/1X & 2 & 16.99 & 37.43 & 44.89 & 46.33 & 35.01 & 21.54 & 73.2 & 59.28 & 68.65 & 22.7 & 42.6 \\
\rowcolor[gray]{0.9}
SamMed 2D & 1PPS + 1PPS Refine & 1X/1X & 3 & 16.57 & 37.5 & 44.52 & 46.4 & 34.15 & 21.76 & 75.4 & 60.96 & 69.23 & 22.01 & 42.85 \\
\rowcolor[gray]{0.9}
SamMed 2D & 1PPS + 1PPS Refine & 1X/1X & 4 & 16.05 & 37.41 & 44.22 & 46.02 & 33.28 & 21.9 & 76.63 & 62.01 & 69.57 & 21.43 & 42.85 \\
\rowcolor[gray]{0.9}
SamMed 2D & 1PPS + 1PPS Refine & 1X/1X & 5 & 15.78 & 37.2 & 43.95 & 45.55 & 32.82 & 21.93 & 77.34 & 62.85 & 69.78 & 21.03 & 42.82 \\
\rowcolor[gray]{0.85}
ScribblePrompt & 1PPS + 1PPS Refine & 1X/1X & 0 & 25.29 & 32.9 & 47.38 & 49.09 & 26.43 & 4.37 & 49.73 & 45.4 & 55.47 & 35.12 & 37.12 \\
\rowcolor[gray]{0.85}
ScribblePrompt & 1PPS + 1PPS Refine & 1X/1X & 1 & 4.85 & 14.18 & 22.53 & 20.7 & 1.79 & 0.47 & 20.28 & 13.97 & 32.68 & 16.44 & 14.79 \\
\rowcolor[gray]{0.85}
ScribblePrompt & 1PPS + 1PPS Refine & 1X/1X & 2 & 11.69 & 21.84 & 28.48 & 25.57 & 16.35 & 2.98 & 34.11 & 31.91 & 46.36 & 26.93 & 24.62 \\
\rowcolor[gray]{0.85}
ScribblePrompt & 1PPS + 1PPS Refine & 1X/1X & 3 & 6.4 & 16.48 & 23.47 & 23.54 & 6.1 & 0.72 & 22.08 & 17.91 & 33.56 & 19.55 & 16.98 \\
\rowcolor[gray]{0.85}
ScribblePrompt & 1PPS + 1PPS Refine & 1X/1X & 4 & 11.75 & 21.51 & 28.54 & 25.55 & 15.73 & 2.99 & 33.37 & 31.02 & 46.18 & 27.24 & 24.39 \\
\rowcolor[gray]{0.85}
ScribblePrompt & 1PPS + 1PPS Refine & 1X/1X & 5 & 6.68 & 16.52 & 23.32 & 22.21 & 6.69 & 0.91 & 20.91 & 17.86 & 32.23 & 17.91 & 16.52 \\
\rowcolor[gray]{1}
SAM & 1PPS + 1PPS Refine* & 1X/1X & 0 & 1.89 & 3.36 & 3.06 & 4.21 & 1.43 & 0.92 & 11.06 & 15.4 & 28.52 & 4.89 & 7.47 \\
\rowcolor[gray]{1}
SAM & 1PPS + 1PPS Refine* & 1X/1X & 1 & 2.31 & 3.95 & 4.8 & 4.31 & 2.93 & 1.27 & 14.36 & 18.66 & 17.76 & 5.55 & 7.59 \\
\rowcolor[gray]{1}
SAM & 1PPS + 1PPS Refine* & 1X/1X & 2 & 2.96 & 5.1 & 5.88 & 6.83 & 3.63 & 1.68 & 17.71 & 22.64 & 25.27 & 6.72 & 9.84 \\
\rowcolor[gray]{1}
SAM & 1PPS + 1PPS Refine* & 1X/1X & 3 & 3.67 & 4.76 & 6.7 & 8.48 & 3.97 & 1.88 & 21.2 & 24.34 & 29.31 & 7.19 & 11.15 \\
\rowcolor[gray]{1}
SAM & 1PPS + 1PPS Refine* & 1X/1X & 4 & 3.84 & 4.27 & 5.99 & 10.49 & 4.77 & 1.85 & 22.79 & 25.3 & 32.39 & 7.33 & 11.9 \\
\rowcolor[gray]{1}
SAM & 1PPS + 1PPS Refine* & 1X/1X & 5 & 2.8 & 4.09 & 4.64 & 12.52 & 5.79 & 1.88 & 24.89 & 25.93 & 32.2 & 7.09 & 12.18 \\
\rowcolor[gray]{0.95}
SAM2 & 1PPS + 1PPS Refine* & 1X/1X & 0 & 3.36 & 4.99 & 4.41 & 5.91 & 3.77 & 1.62 & 15.64 & 22.36 & 45.86 & 6.79 & 11.47 \\
\rowcolor[gray]{0.95}
SAM2 & 1PPS + 1PPS Refine* & 1X/1X & 1 & 5.86 & 13.4 & 13.61 & 13.8 & 9.17 & 4.58 & 43.18 & 42.1 & 60.63 & 12.8 & 21.91 \\
\rowcolor[gray]{0.95}
SAM2 & 1PPS + 1PPS Refine* & 1X/1X & 2 & 10.11 & 20.08 & 19.99 & 20.62 & 14.69 & 8.49 & 59.67 & 51.88 & 70.53 & 18.52 & 29.46 \\
\rowcolor[gray]{0.95}
SAM2 & 1PPS + 1PPS Refine* & 1X/1X & 3 & 14.5 & 25.49 & 25.19 & 26.3 & 19.94 & 11.6 & 66.94 & 58.3 & 75.98 & 23.44 & 34.77 \\
\rowcolor[gray]{0.95}
SAM2 & 1PPS + 1PPS Refine* & 1X/1X & 4 & 19.3 & 30.35 & 28.84 & 30.21 & 25.06 & 14.15 & 72.21 & 87.37 & 79.46 & 27.32 & 41.43 \\
\rowcolor[gray]{0.95}
SAM2 & 1PPS + 1PPS Refine* & 1X/1X & 5 & 23.71 & 34.35 & 32.83 & 33.5 & 29.61 & 15.81 & 74.65 & 90.05 & 81.96 & 30.76 & 44.72 \\
\rowcolor[gray]{0.9}
SamMed 2D & 1PPS + 1PPS Refine* & 1X/1X & 0 & 18.28 & 33.98 & 44.01 & 41.44 & 35.97 & 18.79 & 59.09 & 49.33 & 62.66 & 24.28 & 38.78 \\
\rowcolor[gray]{0.9}
SamMed 2D & 1PPS + 1PPS Refine* & 1X/1X & 1 & 18.65 & 37.46 & 44.63 & 45.02 & 37.47 & 20.36 & 68.45 & 56.05 & 67.19 & 25.01 & 42.03 \\
\rowcolor[gray]{0.9}
SamMed 2D & 1PPS + 1PPS Refine* & 1X/1X & 2 & 18.93 & 39.03 & 45.11 & 46.16 & 37.44 & 20.91 & 72.76 & 59.35 & 69.13 & 25.53 & 43.44 \\
\rowcolor[gray]{0.9}
SamMed 2D & 1PPS + 1PPS Refine* & 1X/1X & 3 & 19.25 & 40.03 & 45.77 & 46.89 & 37.48 & 21.31 & 75.23 & 61.37 & 70.4 & 26.03 & 44.37 \\
\rowcolor[gray]{0.9}
SamMed 2D & 1PPS + 1PPS Refine* & 1X/1X & 4 & 19.6 & 40.85 & 46.52 & 47.45 & 37.68 & 21.69 & 76.85 & 62.81 & 71.34 & 26.5 & 45.13 \\
\rowcolor[gray]{0.9}
SamMed 2D & 1PPS + 1PPS Refine* & 1X/1X & 5 & 19.89 & 41.53 & 47.25 & 47.84 & 37.96 & 22.03 & 77.88 & 63.9 & 72.08 & 26.93 & 45.73 \\
\rowcolor[gray]{0.85}
ScribblePrompt & 1PPS + 1PPS Refine* & 1X/1X & 0 & 25.29 & 32.9 & 47.38 & 49.09 & 26.43 & 4.37 & 49.73 & 45.4 & 55.47 & 35.12 & 37.12 \\
\rowcolor[gray]{0.85}
ScribblePrompt & 1PPS + 1PPS Refine* & 1X/1X & 1 & 28.79 & 36.93 & 55.37 & 56.62 & 30.29 & 7.0 & 51.12 & 38.99 & 64.36 & 41.32 & 41.08 \\
\rowcolor[gray]{0.85}
ScribblePrompt & 1PPS + 1PPS Refine* & 1X/1X & 2 & 35.33 & 42.85 & 62.06 & 66.91 & 39.29 & 11.94 & 59.68 & 44.7 & 69.95 & 45.81 & 47.85 \\
\rowcolor[gray]{0.85}
ScribblePrompt & 1PPS + 1PPS Refine* & 1X/1X & 3 & 39.56 & 47.12 & 65.91 & 71.03 & 43.54 & 18.35 & 64.93 & 50.5 & 72.84 & 48.29 & 52.21 \\
\rowcolor[gray]{0.85}
ScribblePrompt & 1PPS + 1PPS Refine* & 1X/1X & 4 & 41.75 & 49.85 & 67.77 & 74.24 & 46.29 & 26.52 & 70.31 & 55.53 & 74.36 & 49.92 & 55.65 \\
\rowcolor[gray]{0.85}
ScribblePrompt & 1PPS + 1PPS Refine* & 1X/1X & 5 & 43.34 & 51.55 & 68.72 & 75.34 & 48.11 & 34.68 & 73.27 & 58.74 & 75.14 & 50.79 & 57.97 \\
\rowcolor[gray]{1}
SAM & 1PPS + Scribble Refine & 1x/3 & 0 & 1.89 & 3.36 & 3.06 & 4.21 & 1.43 & 0.92 & 11.06 & 15.4 & 28.52 & 4.89 & 7.47 \\
\rowcolor[gray]{1}
SAM & 1PPS + Scribble Refine & 1x/3 & 1 & 0.38 & 4.5 & 1.17 & 8.51 & 1.21 & 0.24 & 9.78 & 9.87 & 24.8 & 6.21 & 6.67 \\
\rowcolor[gray]{1}
SAM & 1PPS + Scribble Refine & 1x/3 & 2 & 0.7 & 3.82 & 2.41 & 12.91 & 1.69 & 0.76 & 9.02 & 10.73 & 26.84 & 5.25 & 7.41 \\
\rowcolor[gray]{1}
SAM & 1PPS + Scribble Refine & 1x/3 & 3 & 0.59 & 4.09 & 1.63 & 6.96 & 1.33 & 0.56 & 8.61 & 10.79 & 27.88 & 6.28 & 6.87 \\
\rowcolor[gray]{1}
SAM & 1PPS + Scribble Refine & 1x/3 & 4 & 0.68 & 3.49 & 1.43 & 8.82 & 0.9 & 0.61 & 9.57 & 10.02 & 25.32 & 5.42 & 6.63 \\
\rowcolor[gray]{1}
SAM & 1PPS + Scribble Refine & 1x/3 & 5 & 0.53 & 3.41 & 1.51 & 7.99 & 1.41 & 0.55 & 7.75 & 10.07 & 26.61 & 6.69 & 6.65 \\
\rowcolor[gray]{0.95}
SAM2 & 1PPS + Scribble Refine & 1x/3 & 0 & 3.36 & 4.99 & 4.41 & 5.91 & 3.77 & 1.62 & 15.64 & 22.36 & 45.86 & 6.79 & 11.47 \\
\rowcolor[gray]{0.95}
SAM2 & 1PPS + Scribble Refine & 1x/3 & 1 & 3.64 & 5.89 & 5.5 & 6.05 & 3.84 & 2.12 & 17.01 & 21.49 & 45.67 & 7.88 & 11.91 \\
\rowcolor[gray]{0.95}
SAM2 & 1PPS + Scribble Refine & 1x/3 & 2 & 3.65 & 6.07 & 5.7 & 5.63 & 2.91 & 2.14 & 17.8 & 20.17 & 43.8 & 7.78 & 11.57 \\
\rowcolor[gray]{0.95}
SAM2 & 1PPS + Scribble Refine & 1x/3 & 3 & 3.45 & 5.89 & 5.27 & 6.32 & 2.71 & 2.22 & 18.64 & 18.56 & 41.95 & 7.64 & 11.27 \\
\rowcolor[gray]{0.95}
SAM2 & 1PPS + Scribble Refine & 1x/3 & 4 & 3.4 & 5.57 & 4.88 & 6.02 & 2.58 & 2.13 & 19.27 & 17.91 & 41.23 & 7.5 & 11.05 \\
\rowcolor[gray]{0.95}
SAM2 & 1PPS + Scribble Refine & 1x/3 & 5 & 3.38 & 5.41 & 4.94 & 5.52 & 2.34 & 1.77 & 18.69 & 18.46 & 40.69 & 7.45 & 10.86 \\
\rowcolor[gray]{0.9}
SamMed 2D & 1PPS + Scribble Refine & 1x/3 & 0 & 18.28 & 33.98 & 44.01 & 41.44 & 35.97 & 18.79 & 59.09 & 49.33 & 62.66 & 24.28 & 38.78 \\
\rowcolor[gray]{0.9}
SamMed 2D & 1PPS + Scribble Refine & 1x/3 & 1 & 17.38 & 35.9 & 44.18 & 44.43 & 34.82 & 18.99 & 67.64 & 55.93 & 66.38 & 23.62 & 40.93 \\
\rowcolor[gray]{0.9}
SamMed 2D & 1PPS + Scribble Refine & 1x/3 & 2 & 16.42 & 35.5 & 43.49 & 43.96 & 32.82 & 19.22 & 72.01 & 57.86 & 66.72 & 22.85 & 41.08 \\
\rowcolor[gray]{0.9}
SamMed 2D & 1PPS + Scribble Refine & 1x/3 & 3 & 16.04 & 35.15 & 42.47 & 43.96 & 31.55 & 18.67 & 71.25 & 58.44 & 66.48 & 22.0 & 40.6 \\
\rowcolor[gray]{0.9}
SamMed 2D & 1PPS + Scribble Refine & 1x/3 & 4 & 15.38 & 34.65 & 42.06 & 43.88 & 29.83 & 18.88 & 71.7 & 58.67 & 66.18 & 21.39 & 40.26 \\
\rowcolor[gray]{0.9}
SamMed 2D & 1PPS + Scribble Refine & 1x/3 & 5 & 15.04 & 34.22 & 41.44 & 43.88 & 28.87 & 19.59 & 71.46 & 59.17 & 65.89 & 20.74 & 40.03 \\
\rowcolor[gray]{0.85}
ScribblePrompt & 1PPS + Scribble Refine & 1x/3 & 0 & 25.29 & 32.9 & 47.37 & 49.09 & 26.43 & 4.37 & 49.73 & 45.4 & 55.47 & 35.12 & 37.12 \\
\rowcolor[gray]{0.85}
ScribblePrompt & 1PPS + Scribble Refine & 1x/3 & 1 & 6.3 & 22.19 & 38.63 & 39.72 & 10.24 & 1.16 & 36.22 & 32.16 & 52.76 & 24.5 & 26.39 \\
\rowcolor[gray]{0.85}
ScribblePrompt & 1PPS + Scribble Refine & 1x/3 & 2 & 13.26 & 24.48 & 38.53 & 33.5 & 13.26 & 2.88 & 36.67 & 31.76 & 51.3 & 32.37 & 27.8 \\
\rowcolor[gray]{0.85}
ScribblePrompt & 1PPS + Scribble Refine & 1x/3 & 3 & 7.99 & 21.89 & 35.94 & 36.36 & 10.49 & 1.21 & 34.76 & 27.56 & 49.24 & 26.1 & 25.15 \\
\rowcolor[gray]{0.85}
ScribblePrompt & 1PPS + Scribble Refine & 1x/3 & 4 & 11.13 & 23.92 & 39.96 & 34.97 & 12.58 & 2.39 & 34.69 & 29.82 & 50.13 & 31.82 & 27.14 \\
\rowcolor[gray]{0.85}
ScribblePrompt & 1PPS + Scribble Refine & 1x/3 & 5 & 7.99 & 20.98 & 36.75 & 32.74 & 9.9 & 1.16 & 33.85 & 31.63 & 48.47 & 27.55 & 25.1 \\
\rowcolor[gray]{1}
SAM & 1PPS + Scribble Refine* & 1X/3 & 0 & 1.89 & 3.36 & 3.06 & 4.21 & 1.43 & 0.92 & 11.06 & 15.4 & 28.52 & 4.89 & 7.47 \\
\rowcolor[gray]{1}
SAM & 1PPS + Scribble Refine* & 1X/3 & 1 & 2.64 & 3.6 & 4.63 & 5.15 & 2.14 & 1.1 & 10.25 & 13.86 & 37.1 & 6.2 & 8.67 \\
\rowcolor[gray]{1}
SAM & 1PPS + Scribble Refine* & 1X/3 & 2 & 2.84 & 4.49 & 5.27 & 8.53 & 2.63 & 1.45 & 11.35 & 15.05 & 39.68 & 6.5 & 9.78 \\
\rowcolor[gray]{1}
SAM & 1PPS + Scribble Refine* & 1X/3 & 3 & 2.97 & 4.3 & 4.74 & 10.05 & 2.48 & 1.63 & 12.45 & 15.15 & 38.94 & 6.1 & 9.88 \\
\rowcolor[gray]{1}
SAM & 1PPS + Scribble Refine* & 1X/3 & 4 & 3.06 & 3.43 & 3.42 & 12.12 & 2.22 & 1.64 & 13.03 & 14.33 & 35.61 & 5.28 & 9.41 \\
\rowcolor[gray]{1}
SAM & 1PPS + Scribble Refine* & 1X/3 & 5 & 2.15 & 2.96 & 2.43 & 11.85 & 2.52 & 1.52 & 13.4 & 14.67 & 33.81 & 4.66 & 9.0 \\
\rowcolor[gray]{0.95}
SAM2 & 1PPS + Scribble Refine* & 1X/3 & 0 & 3.36 & 4.99 & 4.41 & 5.91 & 3.77 & 1.62 & 15.64 & 22.36 & 45.86 & 6.79 & 11.47 \\
\rowcolor[gray]{0.95}
SAM2 & 1PPS + Scribble Refine* & 1X/3 & 1 & 4.88 & 10.1 & 10.19 & 14.09 & 9.61 & 4.0 & 34.23 & 33.66 & 56.19 & 11.36 & 18.83 \\
\rowcolor[gray]{0.95}
SAM2 & 1PPS + Scribble Refine* & 1X/3 & 2 & 5.7 & 12.98 & 14.7 & 20.06 & 13.52 & 6.56 & 41.64 & 41.38 & 59.46 & 13.97 & 23.0 \\
\rowcolor[gray]{0.95}
SAM2 & 1PPS + Scribble Refine* & 1X/3 & 3 & 6.86 & 14.98 & 18.73 & 21.55 & 15.91 & 8.75 & 49.62 & 46.04 & 61.29 & 15.48 & 25.92 \\
\rowcolor[gray]{0.95}
SAM2 & 1PPS + Scribble Refine* & 1X/3 & 4 & 7.43 & 16.58 & 22.06 & 23.03 & 17.47 & 10.33 & 53.22 & 50.35 & 62.16 & 16.6 & 27.92 \\
\rowcolor[gray]{0.95}
SAM2 & 1PPS + Scribble Refine* & 1X/3 & 5 & 7.86 & 17.84 & 25.14 & 23.8 & 19.11 & 11.97 & 55.59 & 52.72 & 62.95 & 17.77 & 29.47 \\
\rowcolor[gray]{0.9}
SamMed 2D & 1PPS + Scribble Refine* & 1X/3 & 0 & 18.28 & 33.98 & 44.01 & 41.44 & 35.97 & 18.79 & 59.09 & 49.33 & 62.66 & 24.28 & 38.78 \\
\rowcolor[gray]{0.9}
SamMed 2D & 1PPS + Scribble Refine* & 1X/3 & 1 & 18.19 & 36.17 & 43.27 & 43.15 & 35.71 & 18.97 & 66.82 & 54.92 & 66.13 & 24.23 & 40.76 \\
\rowcolor[gray]{0.9}
SamMed 2D & 1PPS + Scribble Refine* & 1X/3 & 2 & 18.09 & 37.12 & 42.97 & 42.87 & 35.33 & 18.99 & 69.63 & 57.03 & 67.39 & 24.0 & 41.34 \\
\rowcolor[gray]{0.9}
SamMed 2D & 1PPS + Scribble Refine* & 1X/3 & 3 & 18.03 & 37.6 & 42.73 & 42.56 & 35.15 & 19.0 & 71.31 & 58.23 & 68.06 & 23.9 & 41.66 \\
\rowcolor[gray]{0.9}
SamMed 2D & 1PPS + Scribble Refine* & 1X/3 & 4 & 18.05 & 37.94 & 42.53 & 42.25 & 34.99 & 19.07 & 72.41 & 59.14 & 68.56 & 23.8 & 41.87 \\
\rowcolor[gray]{0.9}
SamMed 2D & 1PPS + Scribble Refine* & 1X/3 & 5 & 18.08 & 38.18 & 42.48 & 42.05 & 34.88 & 19.08 & 73.18 & 59.9 & 68.93 & 23.72 & 42.05 \\
\rowcolor[gray]{0.85}
ScribblePrompt & 1PPS + Scribble Refine* & 1X/3 & 0 & 25.29 & 32.9 & 47.37 & 49.09 & 26.43 & 4.37 & 49.73 & 45.4 & 55.47 & 35.12 & 37.12 \\
\rowcolor[gray]{0.85}
ScribblePrompt & 1PPS + Scribble Refine* & 1X/3 & 1 & 27.54 & 33.07 & 50.97 & 42.56 & 24.39 & 12.87 & 42.75 & 36.19 & 59.75 & 38.81 & 36.89 \\
\rowcolor[gray]{0.85}
ScribblePrompt & 1PPS + Scribble Refine* & 1X/3 & 2 & 32.12 & 35.91 & 52.98 & 50.54 & 27.9 & 20.19 & 45.32 & 37.63 & 61.23 & 41.94 & 40.58 \\
\rowcolor[gray]{0.85}
ScribblePrompt & 1PPS + Scribble Refine* & 1X/3 & 3 & 34.98 & 39.06 & 55.9 & 57.26 & 31.82 & 28.71 & 51.78 & 41.82 & 62.79 & 43.86 & 44.8 \\
\rowcolor[gray]{0.85}
ScribblePrompt & 1PPS + Scribble Refine* & 1X/3 & 4 & 36.65 & 41.64 & 57.53 & 62.96 & 35.13 & 34.67 & 57.44 & 45.05 & 64.59 & 45.99 & 48.17 \\
\rowcolor[gray]{0.85}
ScribblePrompt & 1PPS + Scribble Refine* & 1X/3 & 5 & 37.76 & 43.84 & 59.31 & 67.05 & 37.72 & 41.1 & 60.27 & 47.49 & 65.83 & 47.42 & 50.78 \\
\rowcolor[gray]{1}
SAM & 3B Inter + Scribble Refine* & 6/3 & 0 & 73.31 & 60.53 & 80.06 & NaN & 75.75 & 77.85 & 86.54 & 76.3 & 62.48 & 67.48 & 73.37 \\
\rowcolor[gray]{1}
SAM & 3B Inter + Scribble Refine* & 6/3 & 1 & NaN & 57.97 & 76.5 & NaN & 72.43 & 75.4 & 82.98 & 72.22 & 63.08 & 69.85 & 71.3 \\
\rowcolor[gray]{1}
SAM & 3B Inter + Scribble Refine* & 6/3 & 2 & NaN & 56.94 & 74.58 & NaN & 72.64 & 75.02 & 80.92 & 72.78 & 65.59 & 71.25 & 71.21 \\
\rowcolor[gray]{1}
SAM & 3B Inter + Scribble Refine* & 6/3 & 3 & NaN & 56.17 & 73.97 & NaN & 73.29 & 74.03 & 81.56 & 72.53 & 67.25 & 71.56 & 71.3 \\
\rowcolor[gray]{1}
SAM & 3B Inter + Scribble Refine* & 6/3 & 4 & NaN & 55.13 & 71.9 & NaN & 73.41 & 72.74 & 82.01 & 70.69 & 68.39 & 71.67 & 70.74 \\
\rowcolor[gray]{1}
SAM & 3B Inter + Scribble Refine* & 6/3 & 5 & NaN & 53.44 & 69.1 & NaN & 72.93 & 70.31 & 82.46 & 69.29 & 69.31 & 70.93 & 69.72 \\
\rowcolor[gray]{0.95}
SAM2 & 3B Inter + Scribble Refine* & 6/3 & 0 & 77.29 & 65.06 & 80.2 & 82.62 & 78.6 & 80.43 & 86.24 & 78.41 & 58.8 & 71.4 & 75.9 \\
\rowcolor[gray]{0.95}
SAM2 & 3B Inter + Scribble Refine* & 6/3 & 1 & 77.9 & 66.22 & 82.03 & 84.28 & 79.08 & 81.23 & 86.87 & 80.31 & 61.82 & 71.5 & 77.12 \\
\rowcolor[gray]{0.95}
SAM2 & 3B Inter + Scribble Refine* & 6/3 & 2 & 79.22 & 67.37 & 82.98 & 85.49 & 80.69 & 83.6 & 87.5 & 81.39 & 63.31 & 73.8 & 78.53 \\
\rowcolor[gray]{0.95}
SAM2 & 3B Inter + Scribble Refine* & 6/3 & 3 & 79.18 & 68.25 & 83.8 & 86.33 & 81.22 & 84.11 & 87.74 & 82.12 & 64.37 & 74.96 & 79.21 \\
\rowcolor[gray]{0.95}
SAM2 & 3B Inter + Scribble Refine* & 6/3 & 4 & 78.96 & 69.11 & 84.37 & 86.96 & 81.63 & 84.97 & 87.83 & 82.39 & 65.1 & 76.17 & 79.75 \\
\rowcolor[gray]{0.95}
SAM2 & 3B Inter + Scribble Refine* & 6/3 & 5 & 78.72 & 69.99 & 84.87 & 87.35 & 82.14 & 85.18 & 88.08 & 82.76 & 65.83 & 76.98 & 80.19 \\
\rowcolor[gray]{0.9}
SamMed 2D & 3B Inter + Scribble Refine* & 6/3 & 0 & 39.98 & 52.3 & 75.68 & 68.33 & 56.34 & 56.67 & 80.76 & 69.65 & 58.4 & 44.42 & 60.25 \\
\rowcolor[gray]{0.9}
SamMed 2D & 3B Inter + Scribble Refine* & 6/3 & 1 & 48.17 & 55.46 & 76.12 & 70.87 & 59.0 & 60.15 & 81.76 & 72.27 & 62.02 & 47.75 & 63.36 \\
\rowcolor[gray]{0.9}
SamMed 2D & 3B Inter + Scribble Refine* & 6/3 & 2 & 49.05 & 56.6 & 76.34 & 71.95 & 59.68 & 61.46 & 82.63 & 73.75 & 64.84 & 48.59 & 64.49 \\
\rowcolor[gray]{0.9}
SamMed 2D & 3B Inter + Scribble Refine* & 6/3 & 3 & 48.95 & 57.26 & 76.74 & 72.75 & 60.02 & 61.75 & 83.36 & 74.79 & 67.15 & 49.09 & 65.18 \\
\rowcolor[gray]{0.9}
SamMed 2D & 3B Inter + Scribble Refine* & 6/3 & 4 & 48.89 & 57.75 & 77.04 & 73.15 & 60.28 & 61.94 & 83.92 & 75.54 & 69.02 & 49.41 & 65.7 \\
\rowcolor[gray]{0.9}
SamMed 2D & 3B Inter + Scribble Refine* & 6/3 & 5 & 48.58 & 58.11 & 77.28 & 73.48 & 60.43 & 62.18 & 84.36 & 76.2 & 70.5 & 49.57 & 66.07 \\
\bottomrule
\end{tabular}
}
\caption{\textbf{Interactive refinement results for 2D models across 5 iterations.} The initial prediction is made either using a single point per slice or one of our proposed prompting schemes. If the previous point prompt is reused during refinement, indicated by a *,  the performance increases. The unrealistic slice-wise refinement (1 interaction per slice) is only slightly better than our proposed scribble refinement method (3 interactions).}

%% file: results/interactive_3D.tex
\centering
\resizebox{0.99\linewidth}{!}{
\begin{tabular}{lllllrrrrrrrrrr}
\toprule
Prompter & Model & Interactions & Iteration &D1 & D2 & D3 & D4 & D5 & D6 & D7 & D8 & D9 & D10 & Average \\
\midrule
\rowcolor[gray]{1}
SamMed 3D & 1 center PPV + 1 PPV Refine & 1/1 & 0 & 2.03 & 12.15 & 24.06 & 27.16 & 15.09 & 19.64 & 72.67 & 63.96 & 50.39 & 26.51 & 31.37 \\
\rowcolor[gray]{1}
SamMed 3D & 1 center PPV + 1 PPV Refine & 1/1 & 1 & 3.16 & 13.08 & 27.58 & 36.18 & 14.07 & 23.66 & 72.67 & 78.47 & 49.21 & 26.72 & 34.48 \\
\rowcolor[gray]{1}
SamMed 3D & 1 center PPV + 1 PPV Refine & 1/1 & 2 & 4.12 & 12.99 & 28.43 & 35.44 & 12.5 & 24.17 & 72.87 & 82.11 & 48.78 & 26.71 & 34.81 \\
\rowcolor[gray]{1}
SamMed 3D & 1 center PPV + 1 PPV Refine & 1/1 & 3 & 4.56 & 12.97 & 27.94 & 37.57 & 11.37 & 24.73 & 72.54 & 82.5 & 49.5 & 26.54 & 35.02 \\
\rowcolor[gray]{1}
SamMed 3D & 1 center PPV + 1 PPV Refine & 1/1 & 4 & 4.89 & 13.11 & 27.61 & 34.95 & 10.74 & 24.62 & 72.76 & 82.87 & 49.32 & 27.01 & 34.79 \\
\rowcolor[gray]{1}
SamMed 3D & 1 center PPV + 1 PPV Refine & 1/1 & 5 & 4.94 & 13.04 & 27.34 & 35.73 & 10.34 & 24.77 & 73.33 & 82.92 & 49.46 & 26.95 & 34.88 \\
\rowcolor[gray]{0.95}
SamMed 3D Turbo & 1 center PPV + 1 PPV Refine & 1/1 & 0 & 5.12 & 27.34 & 46.07 & 34.33 & 15.9 & 46.38 & 82.98 & 59.37 & 63.83 & 26.98 & 40.83 \\
\rowcolor[gray]{0.95}
SamMed 3D Turbo & 1 center PPV + 1 PPV Refine & 1/1 & 1 & 5.66 & 28.81 & 48.05 & 37.87 & 16.41 & 50.18 & 86.39 & 73.12 & 67.78 & 28.59 & 44.28 \\
\rowcolor[gray]{0.95}
SamMed 3D Turbo & 1 center PPV + 1 PPV Refine & 1/1 & 2 & 5.82 & 29.45 & 48.7 & 43.93 & 16.94 & 51.88 & 87.33 & 79.33 & 69.55 & 29.79 & 46.27 \\
\rowcolor[gray]{0.95}
SamMed 3D Turbo & 1 center PPV + 1 PPV Refine & 1/1 & 3 & 5.86 & 30.04 & 48.49 & 47.16 & 17.52 & 53.0 & 87.68 & 80.79 & 70.34 & 30.81 & 47.17 \\
\rowcolor[gray]{0.95}
SamMed 3D Turbo & 1 center PPV + 1 PPV Refine & 1/1 & 4 & 5.97 & 30.56 & 48.82 & 48.17 & 18.32 & 54.09 & 87.75 & 82.91 & 71.23 & 31.6 & 47.94 \\
\rowcolor[gray]{0.95}
SamMed 3D Turbo & 1 center PPV + 1 PPV Refine & 1/1 & 5 & 6.18 & 30.92 & 49.61 & 50.09 & 19.22 & 54.68 & 88.17 & 84.43 & 71.93 & 32.13 & 48.73 \\
\rowcolor[gray]{1}
SamMed 3D & 1 center PPV + Scribble Refine & 1/3 & 0 & 2.03 & 12.15 & 24.06 & 27.16 & 15.09 & 19.64 & 72.67 & 63.96 & 50.39 & 26.51 & 31.37 \\
\rowcolor[gray]{1}
SamMed 3D & 1 center PPV + Scribble Refine & 1/3 & 1 & 3.31 & 12.58 & 25.17 & 32.79 & 13.57 & 23.05 & 71.29 & 68.61 & 47.66 & 26.03 & 32.41 \\
\rowcolor[gray]{1}
SamMed 3D & 1 center PPV + Scribble Refine & 1/3 & 2 & 4.04 & 12.93 & 25.91 & 35.22 & 11.66 & 24.54 & 72.38 & 78.9 & 47.65 & 26.23 & 33.95 \\
\rowcolor[gray]{1}
SamMed 3D & 1 center PPV + Scribble Refine & 1/3 & 3 & 4.44 & 12.93 & 25.95 & 36.58 & 10.9 & 24.88 & 73.75 & 84.12 & 48.87 & 26.02 & 34.84 \\
\rowcolor[gray]{1}
SamMed 3D & 1 center PPV + Scribble Refine & 1/3 & 4 & 4.38 & 13.05 & 26.49 & 36.33 & 10.29 & 25.35 & 73.91 & 85.34 & 49.15 & 25.78 & 35.01 \\
\rowcolor[gray]{1}
SamMed 3D & 1 center PPV + Scribble Refine & 1/3 & 5 & 4.55 & 13.23 & 26.83 & 37.68 & 9.85 & 25.8 & 73.67 & 86.4 & 49.65 & 25.91 & 35.36 \\
\rowcolor[gray]{0.95}
SamMed 3D Turbo & 1 center PPV + Scribble Refine & 1/3 & 0 & 5.12 & 27.34 & 46.07 & 34.33 & 15.9 & 46.38 & 82.98 & 59.37 & 63.83 & 26.98 & 40.83 \\
\rowcolor[gray]{0.95}
SamMed 3D Turbo & 1 center PPV + Scribble Refine & 1/3 & 1 & 5.41 & 27.71 & 47.68 & 37.93 & 15.93 & 48.98 & 86.18 & 71.33 & 67.78 & 27.72 & 43.66 \\
\rowcolor[gray]{0.95}
SamMed 3D Turbo & 1 center PPV + Scribble Refine & 1/3 & 2 & 4.87 & 28.19 & 47.72 & 40.62 & 16.59 & 50.72 & 87.2 & 76.94 & 69.91 & 28.94 & 45.17 \\
\rowcolor[gray]{0.95}
SamMed 3D Turbo & 1 center PPV + Scribble Refine & 1/3 & 3 & 4.35 & 28.75 & 47.42 & 41.96 & 17.07 & 52.24 & 87.71 & 78.53 & 70.83 & 30.08 & 45.89 \\
\rowcolor[gray]{0.95}
SamMed 3D Turbo & 1 center PPV + Scribble Refine & 1/3 & 4 & 4.23 & 29.41 & 48.1 & 43.95 & 17.65 & 53.4 & 88.06 & 80.09 & 71.87 & 30.93 & 46.77 \\
\rowcolor[gray]{0.95}
SamMed 3D Turbo & 1 center PPV + Scribble Refine & 1/3 & 5 & 4.32 & 29.84 & 49.16 & 45.71 & 17.89 & 54.63 & 88.24 & 80.88 & 72.5 & 31.55 & 47.47 \\
\bottomrule
\end{tabular}}
    \caption{\textbf{Interactive refinement results for 3D models over 5 iterations.} The initial interaction always starts from a central point of the target object, and refinement is performed either by randomly sampling positive or negative points (1 interaction) or by selecting a point using the proposed scribble refinement method. Scribble drawing is counted as three interactions. In contrast to 2D models, including the previous point prompt did not improve the performance.}
    \label{tab:iterative3d}

%% file: results/organ_vs_patho.tex
\centering
\resizebox{\linewidth}{!}{
\begin{tabular}{llllrrrrrrrrrrrrrr}
\toprule
Prompter & Model & Interactions & D1 & D10 & D2 & D3 & D4 & D5 & D6 & D7 & D8a & D8b & D9 & Average & Path. Average & Org. Average \\
\midrule
\rowcolor[gray]{1}
SAM & 3B Inter & 6 & 75.41 & 67.48 & 60.53 & 80.06 & 81.65 & 75.75 & 77.85 & 86.54 & 73.51 & 79.08 & 62.48 & 74.58 & 79.48 & 67.17 \\
\rowcolor[gray]{0.95}
SAM2 & 3B Inter & 6 & 77.29 & 71.41 & 65.06 & 80.2 & 82.63 & 78.6 & 80.43 & 86.29 & 76.64 & 80.22 & 58.8 & 76.14 & 80.81 & 71.04 \\
\rowcolor[gray]{0.9}
SamMed 2D & 3B Inter & 6 & 39.98 & 44.42 & 52.3 & 75.68 & 68.33 & 56.34 & 56.67 & 80.76 & 68.91 & 70.39 & 58.4 & 61.11 & 64.02 & 55.21 \\
\rowcolor[gray]{0.85}
ScribblePrompt & 3B Inter & 6 & 14.16 & 7.09 & 24.84 & 44.69 & 54.56 & 18.72 & 9.78 & 17.79 & 8.24 & 10.25 & 32.14 & 22.02 & 24.28 & 13.39 \\
\rowcolor[gray]{0.8}
MedSam & 3B Inter & 6 & 57.46 & 49.05 & 60.45 & 61.44 & 65.3 & 64.57 & 61.43 & 72.82 & 66.6 & 68.6 & 38.8 & 60.59 & 64.52 & 58.7 \\
\rowcolor[gray]{1}
SamMed 3D & 1 center PPV & 1 & 2.07 & 26.51 & 12.15 & 24.06 & 27.16 & 15.11 & 19.64 & 72.66 & 63.95 & 42.53 & 50.38 & 32.38 & 29.03 & 34.2 \\
\rowcolor[gray]{0.95}
SamMed 3D Turbo & 1 center PPV & 1 & 5.18 & 26.98 & 27.34 & 46.07 & 34.33 & 15.91 & 46.38 & 82.95 & 58.89 & 60.1 & 63.75 & 42.53 & 41.56 & 37.74 \\
\rowcolor[gray]{0.9}
SegVol & 1 center PPV & 1 & 9.96 & 30.73 & 24.91 & 38.49 & 31.36 & 3.17 & 33.92 & 71.01 & 47.82 & 53.52 & 28.21 & 33.92 & 34.49 & 34.49 \\
\rowcolor[gray]{0.9}
SegVol & 3D Box & 3 & 0.55 & 49.45 & 37.17 & 68.11 & 69.72 & 63.21 & 50.13 & 89.95 & 91.07 & 68.89 & 72.13 & 60.03 & 58.65 & 59.23 \\
\bottomrule
\end{tabular}}
\caption{\textbf{Performance comparison for pathologies and organs.} The table presents the average performance for all pathological datasets (D1, D3–D7, D8b) and all healthy organ datasets (D2, D8a, D10). Dataset D8 is split into two parts: D8a, which includes the liver, and D8b, which focuses on liver tumors as a pathological structure. Dataset D9, containing annotations of bone fractures, was excluded as it does not fit neatly into either category of healthy or pathological structures. * SegVol trained on CT images from the D2 HanSeg dataset, while we tested on MRI images from the same patient cohort.}

%% file: results/realistic_static_modalities.tex
\centering
\resizebox{0.8\linewidth}{!}{
\begin{tabular}{llllrrr}
\toprule
Prompter & Model & Interactions & Average & Average MRI & Average CT \\
\midrule
\rowcolor[gray]{1}
SAM & 3B Inter & 6&74.41 & 72.0 & 75.44 \\
\rowcolor[gray]{0.95}
SAM2 & 3B Inter & 6&75.91 & 74.18 & 76.66 \\
\rowcolor[gray]{0.9}
SamMed 2D & 3B Inter & 6&60.25 & 55.99 & 62.08 \\
\rowcolor[gray]{0.85}
ScribblePrompt & 3B Inter & 6&23.3 & 27.9 & 21.33 \\
\rowcolor[gray]{0.8}
MedSam & 3B Inter & 6&59.89 & 59.78 & 59.94 \\
\rowcolor[gray]{1}
SamMed 3D & 1 center PPV & 1&30.3 & 12.76 & 37.81 \\
\rowcolor[gray]{0.95}
SamMed 3D Turbo & 1 center PPV & 1&40.84 & 26.2 & 47.11 \\
\rowcolor[gray]{0.9}
SegVol & 1 center PPV & 1&32.24 & 24.45 & 35.58 \\
\rowcolor[gray]{0.9}
SegVol & 3D Box & 3&58.04 & 35.28 & 67.8 \\
\bottomrule
\end{tabular}}
\caption{\textbf{Averaged CT and MRI performance:} The table shows the DSC for prompting schemes simulating a realistic \textit{Human Effort}, without refinement, averaged over all MRI datasets (D1-3) and over all CT datasets (D4-10). The performance for models trained on medical data is highly dependent on the training data distribution: For example, SegVol was exclusively trained on CT data, leading to subpar performance on MRI datasets.}

%% file: results/uncertainty_boxes.tex
\centering
\resizebox{\linewidth}{!}{
\begin{tabular}{lllrrrrrrrrrrr}
\toprule
Model & Promter & Interactions &D1 & D2 & D3 & D4 & D5 & D6 & D7 & D8 & D9 & D10 & Average \\
\midrule
\rowcolor[gray]{1}
SAM2 & 3B Inter & 6 & 77.29 & 65.06 & 80.2 & 82.63 & 78.6 & 80.43 & 86.29 & 78.43 & 58.8 & 71.41 & 75.91 \\
\rowcolor[gray]{0.95}
SAM2 & 3B Inter $\pm$3 & 6 & 53.09 & 61.92 & 78.34 & 81.21 & 72.72 & 74.33 & 85.57 & 78.02 & 57.95 & 68.58 & 71.17 \\
\rowcolor[gray]{0.9}
SAM2 & 3B Inter $\pm$5 & 6 & 37.56 & 58.32 & 75.91 & 78.54 & 66.7 & 68.47 & 84.71 & 76.93 & 57.0 & 65.38 & 66.95 \\
\bottomrule
\end{tabular}}
\caption{\textbf{Evaluation of SAM2 segmentation performance with box prompts perturbed by random shifts of up to 3 and up to 5 pixels.} The results assess the model's robustness to slight variations in prompt positioning. Even with shifts up to 5 pixels SAM2 outperforms all 3D models}

%% file: main.bbl
\begin{thebibliography}{50}
\providecommand{\natexlab}[1]{#1}
\providecommand{\url}[1]{\texttt{#1}}
\expandafter\ifx\csname urlstyle\endcsname\relax
  \providecommand{\doi}[1]{doi: #1}\else
  \providecommand{\doi}{doi: \begingroup \urlstyle{rm}\Url}\fi

\bibitem[Antonov et~al.(2024)Antonov, Moskalenko, Shepelev, Krapukhin, Soshin, Konushin, and Shakhuro]{antonov2024rclicksrealisticclicksimulation}
Anton Antonov, Andrey Moskalenko, Denis Shepelev, Alexander Krapukhin, Konstantin Soshin, Anton Konushin, and Vlad Shakhuro.
\newblock Rclicks: Realistic click simulation for benchmarking interactive segmentation, 2024.

\bibitem[Benchoufi et~al.(2020)Benchoufi, Matzner-Lober, Molinari, Jannot, and Soyer]{interobserverradiology}
M. Benchoufi, E. Matzner-Lober, N. Molinari, A.-S. Jannot, and P. Soyer.
\newblock Interobserver agreement issues in radiology.
\newblock \emph{Diagnostic and Interventional Imaging}, 101\penalty0 (10):\penalty0 639–641, 2020.

\bibitem[Bui et~al.(2024)Bui, Hoang, Tran, Doretto, Adjeroh, Patel, Choudhary, and Le]{sam3d}
Nhat-Tan Bui, Dinh-Hieu Hoang, Minh-Triet Tran, Gianfranco Doretto, Donald Adjeroh, Brijesh Patel, Arabinda Choudhary, and Ngan Le.
\newblock Sam3d: Segment anything model in volumetric medical images, 2024.

\bibitem[Cheng et~al.(2023)Cheng, Ye, Deng, Chen, Li, Wang, Su, Huang, Chen, Jiang, Sun, He, Zhang, Zhu, and Qiao]{sammed2d}
Junlong Cheng, Jin Ye, Zhongying Deng, Jianpin Chen, Tianbin Li, Haoyu Wang, Yanzhou Su, Ziyan Huang, Jilong Chen, Lei Jiang, Hui Sun, Junjun He, Shaoting Zhang, Min Zhu, and Yu Qiao.
\newblock Sam-med2d, 2023.

\bibitem[Cheng et~al.(2024)Cheng, Fu, Ye, Wang, Li, Wang, Li, Yao, Chen, Li, Su, Zhu, and He]{IMIS}
Junlong Cheng, Bin Fu, Jin Ye, Guoan Wang, Tianbin Li, Haoyu Wang, Ruoyu Li, He Yao, Junren Chen, Jingwen Li, Yanzhou Su, Min Zhu, and Junjun He.
\newblock Interactive medical image segmentation: A benchmark dataset and baseline, 2024.

\bibitem[de~Verdier et~al.(2024)de~Verdier, Saluja, Gagnon, LaBella, Baid, Tahon, Foltyn-Dumitru, Zhang, Alafif, Baig, Chang, and et~al.]{brats}
Maria~Correia de Verdier, Rachit Saluja, Louis Gagnon, Dominic LaBella, Ujjwall Baid, Nourel~Hoda Tahon, Martha Foltyn-Dumitru, Jikai Zhang, Maram Alafif, Saif Baig, Ken Chang, and et al.
\newblock The 2024 brain tumor segmentation (brats) challenge: Glioma segmentation on post-treatment mri, 2024.

\bibitem[Deng et~al.(2023)Deng, Cui, Liu, Yao, Remedios, Bao, Landman, Wheless, Coburn, Wilson, Wang, Zhao, Fogo, Yang, Tang, and Huo]{samdigital}
Ruining Deng, Can Cui, Quan Liu, Tianyuan Yao, Lucas~W. Remedios, Shunxing Bao, Bennett~A. Landman, Lee~E. Wheless, Lori~A. Coburn, Keith~T. Wilson, Yaohong Wang, Shilin Zhao, Agnes~B. Fogo, Haichun Yang, Yucheng Tang, and Yuankai Huo.
\newblock Segment anything model (sam) for digital pathology: Assess zero-shot segmentation on whole slide imaging, 2023.

\bibitem[Dorent et~al.(2024)Dorent, Khajavi, Idris, Ziegler, Somarouthu, Jacene, LaCasce, Deissler, Ehrhardt, Engelson, Fischer, Gu, Handels, Kasai, Kondo, Maier-Hein, Schnabel, Wang, Wang, Wald, Yang, Zhang, Zhang, Pieper, Harris, Kikinis, and Kapur]{lnq2023challenge}
Reuben Dorent, Roya Khajavi, Tagwa Idris, Erik Ziegler, Bhanusupriya Somarouthu, Heather Jacene, Ann LaCasce, Jonathan Deissler, Jan Ehrhardt, Sofija Engelson, Stefan~M. Fischer, Yun Gu, Heinz Handels, Satoshi Kasai, Satoshi Kondo, Klaus Maier-Hein, Julia~A. Schnabel, Guotai Wang, Litingyu Wang, Tassilo Wald, Guang-Zhong Yang, Hanxiao Zhang, Minghui Zhang, Steve Pieper, Gordon Harris, Ron Kikinis, and Tina Kapur.
\newblock Lnq 2023 challenge: Benchmark of weakly-supervised techniques for mediastinal lymph node quantification, 2024.

\bibitem[Dosovitskiy(2020)]{dosovitskiy2020image}
Alexey Dosovitskiy.
\newblock An image is worth 16x16 words: Transformers for image recognition at scale.
\newblock \emph{arXiv preprint arXiv:2010.11929}, 2020.

\bibitem[Dosovitskiy et~al.(2021)Dosovitskiy, Beyer, Kolesnikov, Weissenborn, Zhai, Unterthiner, Dehghani, Minderer, Heigold, Gelly, Uszkoreit, and Houlsby]{transformers}
Alexey Dosovitskiy, Lucas Beyer, Alexander Kolesnikov, Dirk Weissenborn, Xiaohua Zhai, Thomas Unterthiner, Mostafa Dehghani, Matthias Minderer, Georg Heigold, Sylvain Gelly, Jakob Uszkoreit, and Neil Houlsby.
\newblock An image is worth 16x16 words: Transformers for image recognition at scale, 2021.

\bibitem[Du et~al.(2024)Du, Bai, Huang, and Zhao]{segvol}
Yuxin Du, Fan Bai, Tiejun Huang, and Bo Zhao.
\newblock Segvol: Universal and interactive volumetric medical image segmentation, 2024.

\bibitem[Fu et~al.(2014)Fu, Buerba, Long, Blizzard, Lischuk, Haims, and Grauer]{interraterspine}
Michael~C. Fu, Rafael~A. Buerba, William~D. Long, Daniel~J. Blizzard, Andrew~W. Lischuk, Andrew~H. Haims, and Jonathan~N. Grauer.
\newblock Interrater and intrarater agreements of magnetic resonance imaging findings in the lumbar spine: significant variability across degenerative conditions.
\newblock \emph{The Spine Journal}, 14\penalty0 (10):\penalty0 2442–2448, 2014.

\bibitem[Gong et~al.(2023)Gong, Zhong, Ma, Li, Wang, Zhang, Heng, and Dou]{3dsamadapter}
Shizhan Gong, Yuan Zhong, Wenao Ma, Jinpeng Li, Zhao Wang, Jingyang Zhang, Pheng-Ann Heng, and Qi Dou.
\newblock 3dsam-adapter: Holistic adaptation of sam from 2d to 3d for promptable medical image segmentation, 2023.

\bibitem[He et~al.(2023)He, Bao, Li, Stout, Bjornerud, Grant, and Ou]{SAMBenchmarkHe}
Sheng He, Rina Bao, Jingpeng Li, Jeffrey Stout, Atle Bjornerud, P.~Ellen Grant, and Yangming Ou.
\newblock Computer-vision benchmark segment-anything model (sam) in medical images: Accuracy in 12 datasets, 2023.

\bibitem[He et~al.(2024)He, Guo, Tang, Myronenko, Nath, Xu, Yang, Zhao, Simon, Belue, Harmon, Turkbey, Xu, and Li]{vista3d}
Yufan He, Pengfei Guo, Yucheng Tang, Andriy Myronenko, Vishwesh Nath, Ziyue Xu, Dong Yang, Can Zhao, Benjamin Simon, Mason Belue, Stephanie Harmon, Baris Turkbey, Daguang Xu, and Wenqi Li.
\newblock Vista3d: Versatile imaging segmentation and annotation model for 3d computed tomography, 2024.

\bibitem[Hemalatha et~al.(2018)Hemalatha, Thamizhvani, Dhivya, Joseph, Babu, and Chandrasekaran]{Hemalatha18}
R.J. Hemalatha, T.R. Thamizhvani, A.~Josephin~Arockia Dhivya, Josline~Elsa Joseph, Bincy Babu, and R. Chandrasekaran.
\newblock Active contour based segmentation techniques for medical image analysis.
\newblock In \emph{Medical and Biological Image Analysis}, chapter~2. IntechOpen, Rijeka, 2018.

\bibitem[Hesamian et~al.(2019)Hesamian, Jia, He, and Kennedy]{deeplearningahievements}
Mohammad~Hesam Hesamian, Wenjing Jia, Xiangjian He, and Paul Kennedy.
\newblock Deep learning techniques for medical image segmentation: Achievements and challenges.
\newblock \emph{Journal of Digital Imaging}, 32\penalty0 (4):\penalty0 582–596, 2019.

\bibitem[Hu et~al.(2023)Hu, Xia, Ju, and Li]{sammeetsmedicalimages}
Chuanfei Hu, Tianyi Xia, Shenghong Ju, and Xinde Li.
\newblock When sam meets medical images: An investigation of segment anything model (sam) on multi-phase liver tumor segmentation, 2023.

\bibitem[Huang et~al.(2024)Huang, Yang, Liu, Zhou, Chang, Zhou, Chen, Yu, Chen, Chen, Liu, Chi, Hu, Yue, Li, Grau, Fan, Dong, and Ni]{review_sam}
Yuhao Huang, Xin Yang, Lian Liu, Han Zhou, Ao Chang, Xinrui Zhou, Rusi Chen, Junxuan Yu, Jiongquan Chen, Chaoyu Chen, Sijing Liu, Haozhe Chi, Xindi Hu, Kejuan Yue, Lei Li, Vicente Grau, Deng-Ping Fan, Fajin Dong, and Dong Ni.
\newblock Segment anything model for medical images?
\newblock \emph{Medical Image Analysis}, 2024.

\bibitem[Huang et~al.(2023)Huang, Wang, Deng, Ye, Su, Sun, He, Gu, Gu, Zhang, et~al.]{huang2023stu}
Ziyan Huang, Haoyu Wang, Zhongying Deng, Jin Ye, Yanzhou Su, Hui Sun, Junjun He, Yun Gu, Lixu Gu, Shaoting Zhang, et~al.
\newblock Stu-net: Scalable and transferable medical image segmentation models empowered by large-scale supervised pre-training.
\newblock \emph{arXiv preprint arXiv:2304.06716}, 2023.

\bibitem[Isensee et~al.(2020)Isensee, Jaeger, Kohl, Petersen, and Maier-Hein]{nnunet}
Fabian Isensee, Paul~F. Jaeger, Simon A.~A. Kohl, Jens Petersen, and Klaus~H. Maier-Hein.
\newblock nnu-net: a self-configuring method for deep learning-based biomedical image segmentation.
\newblock \emph{Nature Methods}, 18\penalty0 (2):\penalty0 203–211, 2020.

\bibitem[Isensee et~al.(2023)Isensee, Ulrich, Wald, and Maier-Hein]{extending}
Fabian Isensee, Constantin Ulrich, Tassilo Wald, and Klaus~H. Maier-Hein.
\newblock Extending nnu-net is all you need.
\newblock In \emph{Bildverarbeitung f{\"u}r die Medizin 2023}, 2023.

\bibitem[Kirillov et~al.(2023)Kirillov, Mintun, Ravi, Mao, Rolland, Gustafson, Xiao, Whitehead, Berg, Lo, Dollár, and Girshick]{sam}
Alexander Kirillov, Eric Mintun, Nikhila Ravi, Hanzi Mao, Chloe Rolland, Laura Gustafson, Tete Xiao, Spencer Whitehead, Alexander~C. Berg, Wan-Yen Lo, Piotr Dollár, and Ross Girshick.
\newblock Segment anything, 2023.

\bibitem[Li et~al.(2024)Li, Liu, Hu, Wang, and Oguz]{prism}
Hao Li, Han Liu, Dewei Hu, Jiacheng Wang, and Ipek Oguz.
\newblock Prism: A promptable and robust interactive segmentation model with visual prompts, 2024.

\bibitem[Liu et~al.(2023)Liu, Yibulayimu, Sang, Zhu, Wang, Zhao, and Wu]{liu2023pelvic}
Yanzhen Liu, Sutuke Yibulayimu, Yudi Sang, Gang Zhu, Yu Wang, Chunpeng Zhao, and Xinbao Wu.
\newblock Pelvic fracture segmentation using a multi-scale distance-weighted neural network.
\newblock In \emph{International Conference on Medical Image Computing and Computer-Assisted Intervention}, pages 312--321. Springer, 2023.

\bibitem[Luo et~al.(2023)Luo, Fu, Zhong, Liu, Han, Astaraki, Bendazzoli, Toma-Dasu, Ye, Chen, et~al.]{luo2023segrap2023}
Xiangde Luo, Jia Fu, Yunxin Zhong, Shuolin Liu, Bing Han, Mehdi Astaraki, Simone Bendazzoli, Iuliana Toma-Dasu, Yiwen Ye, Ziyang Chen, et~al.
\newblock Segrap2023: A benchmark of organs-at-risk and gross tumor volume segmentation for radiotherapy planning of nasopharyngeal carcinoma.
\newblock \emph{arXiv preprint arXiv:2312.09576}, 2023.

\bibitem[Ma et~al.(2024)Ma, He, Li, Han, You, and Wang]{medsam}
Jun Ma, Yuting He, Feifei Li, Lin Han, Chenyu You, and Bo Wang.
\newblock Segment anything in medical images.
\newblock \emph{Nature Communications}, 15\penalty0 (1), 2024.

\bibitem[Maier-Hein et~al.(2024)Maier-Hein, Reinke, Godau, Tizabi, Buettner, Christodoulou, Glocker, Isensee, Kleesiek, Kozubek, Reyes, and et~al.]{metricsreloaded}
Lena Maier-Hein, Annika Reinke, Patrick Godau, Minu~D. Tizabi, Florian Buettner, Evangelia Christodoulou, Ben Glocker, Fabian Isensee, Jens Kleesiek, Michal Kozubek, Mauricio Reyes, and et al.
\newblock Metrics reloaded: recommendations for image analysis validation.
\newblock \emph{Nature Methods}, 2024.

\bibitem[Marinov et~al.(2024)Marinov, Jäger, Egger, Kleesiek, and Stiefelhagen]{Marinov}
Zdravko Marinov, Paul~F. Jäger, Jan Egger, Jens Kleesiek, and Rainer Stiefelhagen.
\newblock Deep interactive segmentation of medical images: A systematic review and taxonomy.
\newblock \emph{IEEE Transactions on Pattern Analysis and Machine Intelligence}, 46\penalty0 (12):\penalty0 10998--11018, 2024.

\bibitem[Mazurowski et~al.(2023)Mazurowski, Dong, Gu, Yang, Konz, and Zhang]{mazurowski2023segment}
Maciej~A Mazurowski, Haoyu Dong, Hanxue Gu, Jichen Yang, Nicholas Konz, and Yixin Zhang.
\newblock Segment anything model for medical image analysis: an experimental study.
\newblock \emph{Medical Image Analysis}, 2023.

\bibitem[Moawad et~al.(2021)Moawad, Fuentes, Morshid, Khalaf, Elmohr, Abusaif, Hazle, Kaseb, Hassan, Mahvash, Szklaruk, Qayyom, and Elsayes]{hcctace}
Ahmed~W. Moawad, David Fuentes, Ali Morshid, Ahmed~M. Khalaf, Mohab~M. Elmohr, Abdelrahman Abusaif, John~D. Hazle, Ahmed~O. Kaseb, Manal Hassan, Armeen Mahvash, Janio Szklaruk, Aliyya Qayyom, and Khaled Elsayes.
\newblock Multimodality annotated hcc cases with and without advanced imaging segmentation, 2021.

\bibitem[Moawad et~al.(2023)Moawad, Ahmed, ElMohr, Eltaher, Habra, Fisher, Perrier, Zhang, Fuentes, and Elsayes]{adrenalacc}
Ahmed~W. Moawad, Ayahallah~A. Ahmed, Mohab ElMohr, Mohamed Eltaher, Mouhammed~Amir Habra, Sarah Fisher, Nancy Perrier, Miao Zhang, David Fuentes, and Khaled Elsayes.
\newblock Voxel-level segmentation of pathologically-proven adrenocortical carcinoma with ki-67 expression (adrenal-acc-ki67-seg), 2023.

\bibitem[Mohapatra et~al.(2023)Mohapatra, Gosai, and Schlaug]{samvsbet}
Sovesh Mohapatra, Advait Gosai, and Gottfried Schlaug.
\newblock Sam vs bet: A comparative study for brain extraction and segmentation of magnetic resonance images using deep learning, 2023.

\bibitem[Muslim et~al.(2022)Muslim, Mashohor, Gawwam, Mahmud, binti Hanafi, Alnuaimi, Josephine, and Almutairi]{MUSLIM2022108139}
Ali~M. Muslim, Syamsiah Mashohor, Gheyath~Al Gawwam, Rozi Mahmud, Marsyita binti Hanafi, Osama Alnuaimi, Raad Josephine, and Abdullah~Dhaifallah Almutairi.
\newblock Brain mri dataset of multiple sclerosis with consensus manual lesion segmentation and patient meta information.
\newblock \emph{Data in Brief}, 2022.

\bibitem[Podobnik et~al.(2023)Podobnik, Strojan, Peterlin, Ibragimov, and Vrtovec]{podobnik2023han}
Ga{\v{s}}per Podobnik, Primo{\v{z}} Strojan, Primo{\v{z}} Peterlin, Bulat Ibragimov, and Toma{\v{z}} Vrtovec.
\newblock Han-seg: The head and neck organ-at-risk ct and mr segmentation dataset.
\newblock \emph{Medical physics}, 50\penalty0 (3):\penalty0 1917--1927, 2023.

\bibitem[Ravi et~al.(2024)Ravi, Gabeur, Hu, Hu, Ryali, Ma, Khedr, Rädle, Rolland, Gustafson, Mintun, Pan, Alwala, Carion, Wu, Girshick, Dollár, and Feichtenhofer]{sam2}
Nikhila Ravi, Valentin Gabeur, Yuan-Ting Hu, Ronghang Hu, Chaitanya Ryali, Tengyu Ma, Haitham Khedr, Roman Rädle, Chloe Rolland, Laura Gustafson, Eric Mintun, Junting Pan, Kalyan~Vasudev Alwala, Nicolas Carion, Chao-Yuan Wu, Ross Girshick, Piotr Dollár, and Christoph Feichtenhofer.
\newblock Sam 2: Segment anything in images and videos, 2024.

\bibitem[Rokuss et~al.(2025)Rokuss, Kirchhoff, Akbal, Kovacs, Roy, Ulrich, Wald, Rotkopf, Schlemmer, and Maier-Hein]{rokuss2025lesionlocator}
Maximilian Rokuss, Yannick Kirchhoff, Seval Akbal, Balint Kovacs, Saikat Roy, Constantin Ulrich, Tassilo Wald, Lukas~T. Rotkopf, Heinz-Peter Schlemmer, and Klaus Maier-Hein.
\newblock Lesionlocator: Zero-shot universal tumor segmentation and tracking in 3d whole-body imaging, 2025.

\bibitem[Roy et~al.(2023)Roy, Wald, Koehler, Rokuss, Disch, Holzschuh, Zimmerer, and Maier-Hein]{sammd}
Saikat Roy, Tassilo Wald, Gregor Koehler, Maximilian~R. Rokuss, Nico Disch, Julius Holzschuh, David Zimmerer, and Klaus~H. Maier-Hein.
\newblock Sam.md: Zero-shot medical image segmentation capabilities of the segment anything model, 2023.

\bibitem[Ryali et~al.(2023)Ryali, Hu, Bolya, Wei, Fan, Huang, Aggarwal, Chowdhury, Poursaeed, Hoffman, Malik, Li, and Feichtenhofer]{hiera}
Chaitanya Ryali, Yuan-Ting Hu, Daniel Bolya, Chen Wei, Haoqi Fan, Po-Yao Huang, Vaibhav Aggarwal, Arkabandhu Chowdhury, Omid Poursaeed, Judy Hoffman, Jitendra Malik, Yanghao Li, and Christoph Feichtenhofer.
\newblock Hiera: A hierarchical vision transformer without the bells-and-whistles, 2023.

\bibitem[Simpson et~al.(2023)Simpson, Peoples, Creasy, Fichtinger, Gangai, Lasso, Keshava~Murthy, Shia, D'Angelica, and Do]{livermets}
Amber~L. Simpson, Jacob Peoples, John~M. Creasy, Gabor Fichtinger, Natalie Gangai, Andras Lasso, Krishna~Nand Keshava~Murthy, Jinru Shia, Michael~I. D'Angelica, and Richard K.~G. Do.
\newblock Preoperative ct and survival data for patients undergoing resection of colorectal liver metastases (colorectal-liver-metastases), 2023.

\bibitem[Ulrich et~al.(2023)Ulrich, Isensee, Wald, Zenk, Baumgartner, and Maier-Hein]{ulrich2023multitalent}
Constantin Ulrich, Fabian Isensee, Tassilo Wald, Maximilian Zenk, Michael Baumgartner, and Klaus~H Maier-Hein.
\newblock Multitalent: A multi-dataset approach to medical image segmentation.
\newblock In \emph{International Conference on Medical Image Computing and Computer-Assisted Intervention}, pages 648--658. Springer, 2023.

\bibitem[Wahid et~al.(2024)Wahid, Dede, Naser, and Fuller]{hntsmrg2024wahid}
Kareem Wahid, Cem Dede, Mohamed Naser, and Clifton Fuller.
\newblock Training dataset for hntsmrg 2024 challenge, 2024.

\bibitem[Wang et~al.(2024)Wang, Guo, Ye, Deng, Cheng, Li, Chen, Su, Huang, Shen, Fu, Zhang, He, and Qiao]{sammed3d}
Haoyu Wang, Sizheng Guo, Jin Ye, Zhongying Deng, Junlong Cheng, Tianbin Li, Jianpin Chen, Yanzhou Su, Ziyan Huang, Yiqing Shen, Bin Fu, Shaoting Zhang, Junjun He, and Yu Qiao.
\newblock Sam-med3d: Towards general-purpose segmentation models for volumetric medical images, 2024.

\bibitem[Wasserthal et~al.(2023)Wasserthal, Breit, Meyer, Pradella, Hinck, Sauter, Heye, Boll, Cyriac, Yang, et~al.]{wasserthal2023totalsegmentator}
Jakob Wasserthal, Hanns-Christian Breit, Manfred~T Meyer, Maurice Pradella, Daniel Hinck, Alexander~W Sauter, Tobias Heye, Daniel~T Boll, Joshy Cyriac, Shan Yang, et~al.
\newblock Totalsegmentator: robust segmentation of 104 anatomic structures in ct images.
\newblock \emph{Radiology: Artificial Intelligence}, 5\penalty0 (5), 2023.

\bibitem[Wong et~al.(2024)Wong, Rakic, Guttag, and Dalca]{ScribblePrompt}
Hallee~E. Wong, Marianne Rakic, John Guttag, and Adrian~V. Dalca.
\newblock Scribbleprompt: Fast and flexible interactive segmentation for any biomedical image.
\newblock \emph{European Conference on Computer Vision (ECCV)}, 2024.

\bibitem[Wu et~al.(2024)Wu, Zhu, Jin, and Xu]{OnePrompt}
Junde Wu, Jiayuan Zhu, Yueming Jin, and Min Xu.
\newblock One-prompt to segment all medical images, 2024.

\bibitem[Zhang and Liu(2023)]{SAMed}
Kaidong Zhang and Dong Liu.
\newblock Customized segment anything model for medical image segmentation, 2023.

\bibitem[Zhang and Shen(2024)]{zhang2024unleashingpotentialsam2biomedical}
Yichi Zhang and Zhenrong Shen.
\newblock Unleashing the potential of sam2 for biomedical images and videos: A survey, 2024.

\bibitem[Zhao et~al.(2015)Zhao, Schwartz, Kris, and Riely]{rider_lung}
Binsheng Zhao, Lawrence~H Schwartz, Mark~G Kris, and Gregory~J Riely.
\newblock Coffee-break lung ct collection with scan images reconstructed at multiple imaging parameters.
\newblock In \emph{The Cancer Imaging Archive}, 2015.

\bibitem[Zhou et~al.(2023)Zhou, Zhang, Zhou, Wu, and Gong]{samsegmentpolyps}
Tao Zhou, Yizhe Zhang, Yi Zhou, Ye Wu, and Chen Gong.
\newblock Can sam segment polyps?
\newblock \emph{arXiv 2304.07583}, 2023.

\end{thebibliography}
